%% file: main.tex
\documentclass[a4paper,11pt]{article}

\include{preamble}

\begin{document}

\title{Fusion of complementary 2D and 3D mesostructural datasets using generative adversarial networks}


\author[1]{{Amir Dahari}
\thanks{a.dahari@imperial.ac.uk}\ \ }

\author[1]{{Steve Kench}}

\author[1]{{Isaac Squires}}

\author[1]{Samuel J. Cooper \thanks{samuel.cooper@imperial.ac.uk}\ \ }

\affil[1]{{\textit{\footnotesize Dyson School of Design Engineering, Imperial College London, London SW7 2DB}}}

\lhead{\scshape Dahari \textit{et al.}}
\chead{\scshape Image data fusion}
\rhead{\scshape Preprint}

\maketitle

\begin{abstract}
\begin{center}
\begin{minipage}{0.85\textwidth}
{\small  Modelling the impact of a material's mesostructure on device level performance typically requires access to 3D image data containing all the relevant information to define the geometry of the simulation domain. This image data must include sufficient contrast between phases to distinguish each material, be of high enough resolution to capture the key details, but also have a large enough 3D field-of-view to be representative of the material in general. It is rarely possible to obtain data with all of these properties from a single imaging technique. In this paper, we present a method for combining information from pairs of distinct but complementary imaging techniques in order to accurately reconstruct the desired multi-phase, high-resolution, representative, 3D images. Specifically, we use deep convolutional generative adversarial networks to implement super-resolution, style-transfer and dimensionality expansion. Three open-access datasets of energy materials are used to validate the quality of the volumes generated and demonstrate the widespread applicability of the approach, each illustrating a distinct material morphology and imaging approach. Having confidence in the accuracy of our method, based on key mesostructural metrics, we then demonstrate its power using a real data pair from a lithium-ion battery electrode, where the resulting 3D high-resolution image data is not available in the literature. We believe this data-driven approach is superior to previously reported statistical material reconstruction methods, both in terms of its fidelity and ease of use. Furthermore, much of the data required to train this algorithm already exists in the literature, waiting to be combined. As such, our open-source code could precipitate a step change in the computational materials sciences by generating the hard to obtain high quality image volumes necessary to simulate behaviour at the mesoscale.}

\end{minipage}
\end{center}
\end{abstract}
\vspace{.2cm}
\begin{multicols}{2}
\section{Introduction}
\label{Introduction}
The geometrical arrangement of porous or composite materials at the mesoscale can significantly impact the performance of the devices they constitute. In order to accurately model the physical processes that mediate this relationship, it is necessary to have geometric information about the distribution of the various phases. This is typically derived from imaging data, which must have the following four properties: Firstly, it must be three-dimensional (3D), since many percolation networks are not reducible to two dimensions. Secondly, the image data must be of high enough resolution to capture the key details. Thirdly, it must have a large enough field-of-view (FoV) to be representative of the material in general. Finally, it must be possible to confidently differentiate between each of the material phases present. It is rarely possible to obtain data with all four of these properties from a single imaging technique \cite{burnett2019completing}, meaning that the resulting simulation domains may be lacking some vital information, which can undermine the value of any conclusions drawn from these numerical analyses.

Lithium-ion battery cathodes exemplify this kind of challenging imaging scenario. They are usually porous composites where the solid phase is comprised of a ceramic ``active material'' (AM) that is a lithium intercalation compound (e.g. nickel manganese cobalt oxide or NMC), as well as a polymeric binder material containing carbon black called the ``carbon binder domain'' (CBD) which provides both mechanical integrity and electronic conductivity. The pores are filled with a liquid electrolyte and must be well percolated to facilitate transport of the dissolved lithium ions. These electrodes are typically around 100 $\muup$m thick with particle size distributions varying along their thickness; whereas the CBD may have relevant features below 100 nm that critically impact performance. 

Imaging these cathode materials in 3D using X-ray computed tomography (XCT) enables a large, representative field-of-view (FoV) to be observed; however, current XCT technology does not provide sufficient contrast for the CBD phase to be confidently differentiated from the pores \cite{Pietsch2018}. Furthermore, even if the contrast could be enhanced, the resolution would be insufficient (\textit{c.} 100 nm) to observe the CBD's nanoscale features \cite{Lu2020a}. It is also common to image cathode materials with focused ion beam scanning electron microscopy (FIB-SEM), which can provide much higher resolution data (\textit{c.} 10 nm), but the imaged regions are usually quite small as a result. SEM is better than XCT at distinguishing between the solid phases (\textit{i.e.} AM vs. CBD); however, since SEM is based on detecting electrons that have escaped the imaging surface (either backscattered or secondary), it is difficult to distinguish between material that is on the desired imaging plane compared to the inner surface of pores. This ``pore-backs'' problem has several proposed solutions, including infiltration \cite{Lagadec2016} and advanced image tracking \cite{Moroni2020}. A recent paper by the authors describes a strategy to overcome this issue using an \textit{in-situ} platinum deposition approach called \textit{kintsugi} imaging \cite{cooper2022methods}. An additional common limitation of FIB-SEM generated volumes is that uniform spacing in the ``slicing'' direction is hard to guarantee \cite{Jones2014} and so any observed anisotropy may be artificial. 

The 2D SEM imaging technique is, however, far better at differentiating the CBD phase through the use of various contrast enhancement techniques \cite{Kroll2021}. However, as highlighted in Gayon-Lombardo \textit{et al.} \cite{Gayon-Lombardo2020}, for isotropic materials, although the statistical information present in a representative two-dimensional (2D) slice is sufficient to extract basic metrics such as volume fraction and surface area, certain key performance metrics for physics simulation are only available in 3D domains, such as percolation and the tortuosity factor.


To close this gap, our proposed method achieves an accurate reconstruction of high-resolution (high-res), large FoV and multi-phase 3D volumes by fusing together information from complementary 2D and 3D imaging techniques. For this, we use a machine learning framework called generative adversarial networks (GANs)\cite{goodfellow2014generative}. GANs have previously been used to (1) enhance the resolution of (``super-resolve'') colour photographs\cite{ledig2017photo} and (2) transfer style between images (e.g. converting a photograph to a painting)\cite{karras2020analyzing}. Furthermore, Kench and Cooper recently published a GAN architecture able to (3) generate 3D volumes from a single representative 2D slice \cite{kench2021generating} that can be (4) arbitrarily large and therefore representative of the material in general. We propose a GAN architecture that combines these 3 capabilities, designed specifically for the generation of materials mesostructure phase maps. 

Unlike the state-of-the-art in the field of mesostructural generation \cite{niu2021towards}, our approach does not require the user to select and extract characteristic metrics in order to generate or modify their images, but instead it learns the key features and relationships directly from the data. Also, unlike a recent paper\cite{jung2021super} that presents a 2D to 2D super-resolution (super-res) mesostructure technique, our approach not only super-resolves 3D mesostructure, but is also able to add missing phases, change existing ones and capture anisotropy. Further recent related studies can be seen in \cite{furat2022super, fan2021adversarial, jackson2022deep}.

\begin{figure*}
    \centering
    \includegraphics[width=1\textwidth]{ 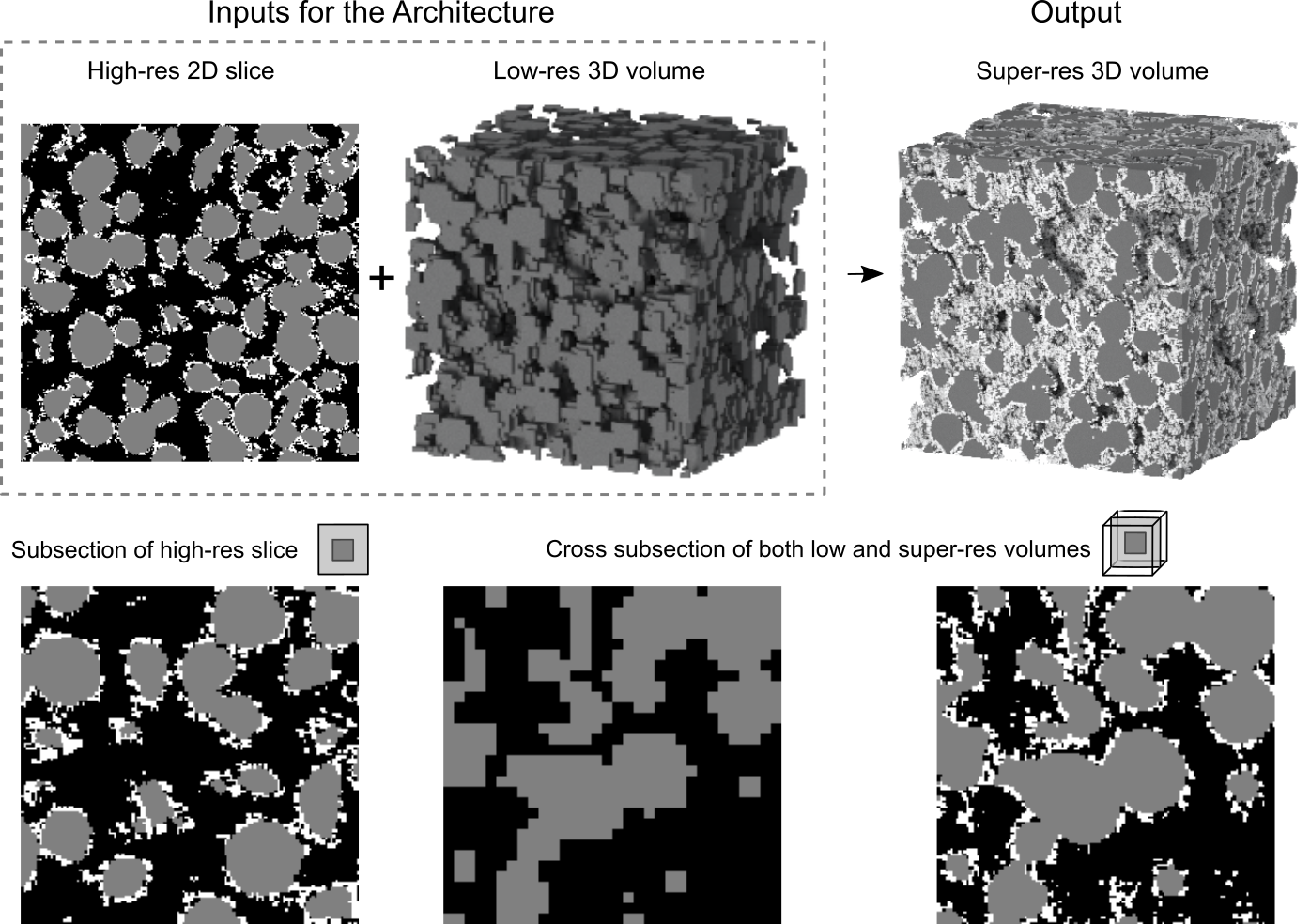}
    \caption{The SuperRes model inputs and output evaluated on the isotropic NMC cathode XCT dataset (Case study 1) \cite{usseglio2018resolving}, with a scale factor of 4. The different phases are pore (black), active material (gray) and binder (white). The top row shows the inputs and output of the model. For training, the model requires a high-res multi-phase 2D image and a low-res binary 3D volume ($\times 4$ stretched to make comparison to the super-res version easier), and the output is a super-res volume of the low-res volume with the same multi-phase characteristics of the high-res 2D image. The bottom row shows a comparison between a random cross section in the same position of the low-res and super-res volumes, and as well as the middle region of the high-res slice to compare the nature of the fine details.}
    \label{fig:Fig1-NMCimages}
\end{figure*}

To illustrate the capability and accuracy of the proposed method, we present four case studies, each based on open-access imaging data. In each study, the paired datasets used for training consist of high-res 2D images and a low-resolution (low-res) 3D volume, with the aim of producing a super resolved (super-res) volume. The first three cases serve to validate the accuracy of the method in distinct scenarios, while the fourth case demonstrates the application of the approach on real paired datasets from the literature. Since the first three case studies require access to the `ground truth' in order to compare the calculated metrics, the low-res volumes were created by blurring then down-sampling the original high-res versions and in some cases removing phases or features to showcase additional functionality. More details about the training data and how it was prepared for analysis can be found in the Supplementary Information Table \ref{tab:Scenarios}. The four case studies are as follows: 

\begin{enumerate}[wide, labelwidth=0pt, labelindent=0pt, label=Case \arabic*:]
    \item Battery cathode (validation on an isotropic material with abundant training data): A low-res, large FoV volume showing only two of the three phases, is combined with a high-res three-phase 2D image to produce a large, 3D, three-phase super-res volume. The material is isotropic and training data is abundant. Detailed statistical comparisons are made to the original 3D high-res volume from which the low-res volume and 2D slice were obtained. In addition, an exploration is made on the impact of the scale factor (the ratio between the scale of the low-res and the corresponding super-res or high-res data) on the reconstruction quality.
    \label{validation}
    \item Battery separator (validation on anisotropic material with limited training data): A low-res, large FoV, two-phase volume is combined with a set of orthogonal high-res 2D images containing fine features to produce a large, 3D, super-res volume. The material is anisotropic and training data is limited. Detailed statistical comparisons are made to the original volume.
    \item Fuel cell anode (validation on isotropic material with abundant training data): To highlight an alternative operational mode of the algorithm, the low-res data in this case is undersampled, rather than underresolved. This also means that all three material phases are represented at both resolutions. 
    \item Battery cathode (demonstration): A low-res, large FoV XCT volume showing only two phases is combined with a high-res, three-phase, 2D SEM image to produce a large, 3D, three-phase, super-res volume. The material is isotropic and training data is limited. The ground truth is not known. To the authors knowledge, the generated volume (c. $2000^3$ voxels, for a 96 \textmu m edge length cubic volume at 50 nm resolution) is the largest three phase Li-ion cathode mesostructural dataset currently available in the literature, which can be downloaded via the link in the code repository.
\end{enumerate}

The mesostructures that are examined in this paper are all battery or fuel cell related, hence the relevance of the mesoscale transport properties that are extracted. However, our model is material and lengthscale agnostic and can be applied to diverse scenarios including catalyst beds, rock formations, or snow and soil packing.

\section{Results}
\label{Results} 

\subsection{Case 1: Validation using an isotropic, three-phase Li-ion battery cathode dataset}

\begin{figure*}[!htb]
    \centering
    \includegraphics[width=\textwidth]{ 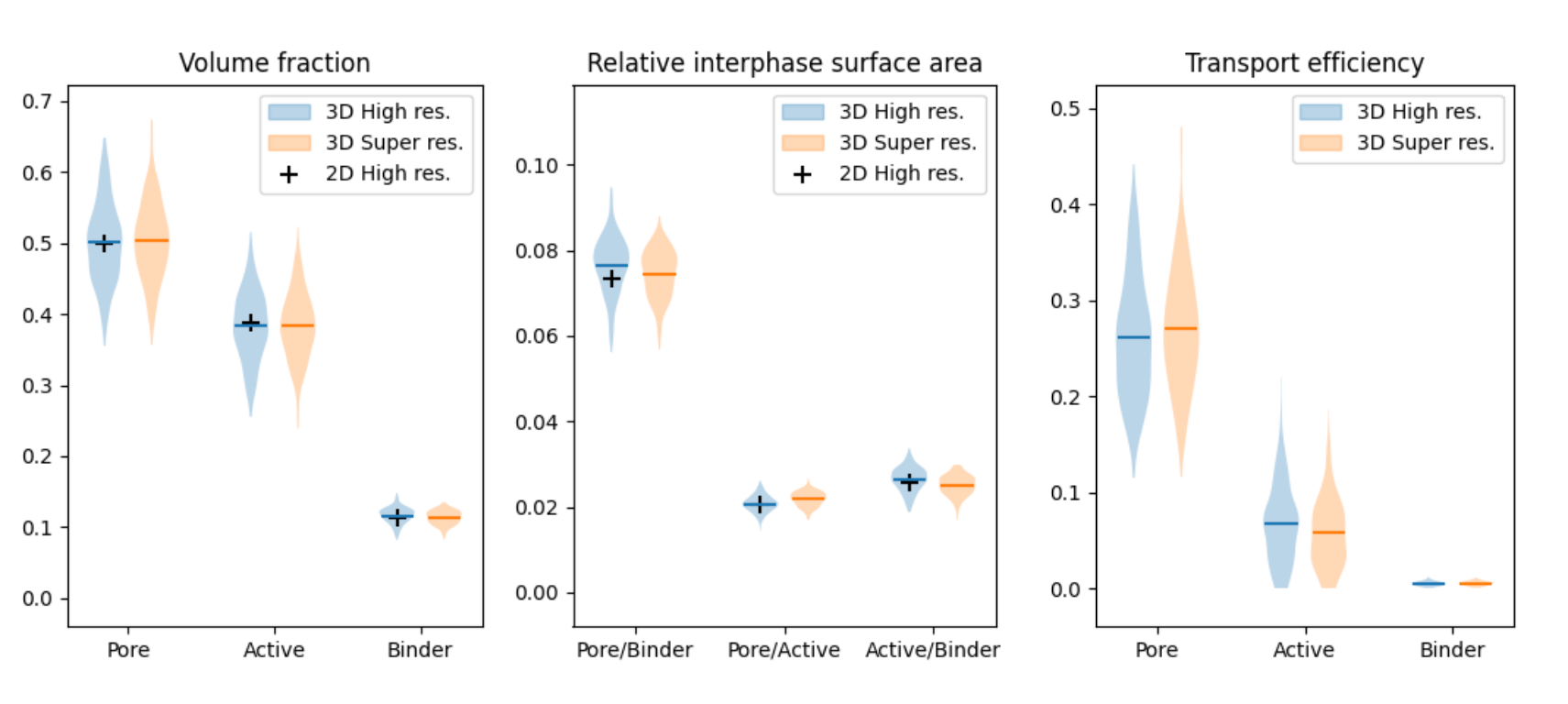}
    \caption{Statistical comparison of key mesostructural metrics between the original high-res 3D volume (blue), the high-res 2D slice (black) and the reconstructed super-res 3D volume (orange). The left figure shows a comparison of the volume fraction of the different phases in the material. The middle figure shows a comparison of the interphase surface area, the fraction of the number of neighboring voxels of different phases by the total number of faces in the volume. The right figure shows a comparison of the transport efficiency for each phase calculated using a diffusivity simulation\cite{cooper2016taufactor}. All metric calculations were obtained using the {\em TauFactor}\cite{cooper2016taufactor} package. All figures show a violin plot distribution of the 3D high-res and the 3D super-res taken from 256 randomly sampled $64^3$ volumes, with normalised width for each distribution and corresponding horizontal lines for the means. The black pluses mark the metrics measured on the whole 2D image. Note that since transport efficiency is a volumetric measurement, only the values for the 2D high-res slice are not included.}
    \label{fig:nmc_metrics}
\end{figure*}

To illustrate the method's capabilities for isotropic, three-phase materials with abundant training data, an open-access dataset from the National Renewable Energy Laboratory (NREL) was used \cite{usseglio2018resolving}. This dataset was derived from XCT imaging of a lithium-ion battery cathode material. The NREL team performed a binary segmentation on the XCT dataset to show the distribution of the active material only and then the binder phase was added stochastically based on statistical information from an SEM image. Full details on the imaging, segmentation, and augmentation of this data can be found in the repository associated with Usseglio-Viretta \textit{et al.}\cite{usseglio2018resolving}. 

For this experiment, in order to have a very large ``ground truth'' volume to showcase the power of the approach, a 3D high-res three-phase volume of $1024^3$ voxels was generated using a {\em SliceGAN} \cite{kench2021generating} trained on the three-phase volume from \cite{usseglio2018resolving}. Next, in order to imitate a dataset representing the same region, but captured using XCT at $4\times$ lower resolution, the following three operations were performed. Firstly, the pore and binder phases were merged together, since the binder phase is typically indistinguishable from the pore phase in XCT due to the very low X-ray attenuation of the binder compared to the active material. Secondly, a Gaussian blur was applied to the active material phase followed by a down sampling by a factor of 4. As described in the methods section, by adjusting the kernel size and standard deviation of this blur operation, this step could be used to simulate ``pixel binning'' all the way through to under sampling. Lastly, the low-res volume was then thresholded to reconvert it into a phase map. This process of creating the simulated low-res volume is described in Supplementary Information Figure \ref{fig:creation_of_low_res_data}.

\begin{figure*}[!htb]
    \centering
    \includegraphics[width=\textwidth]{ 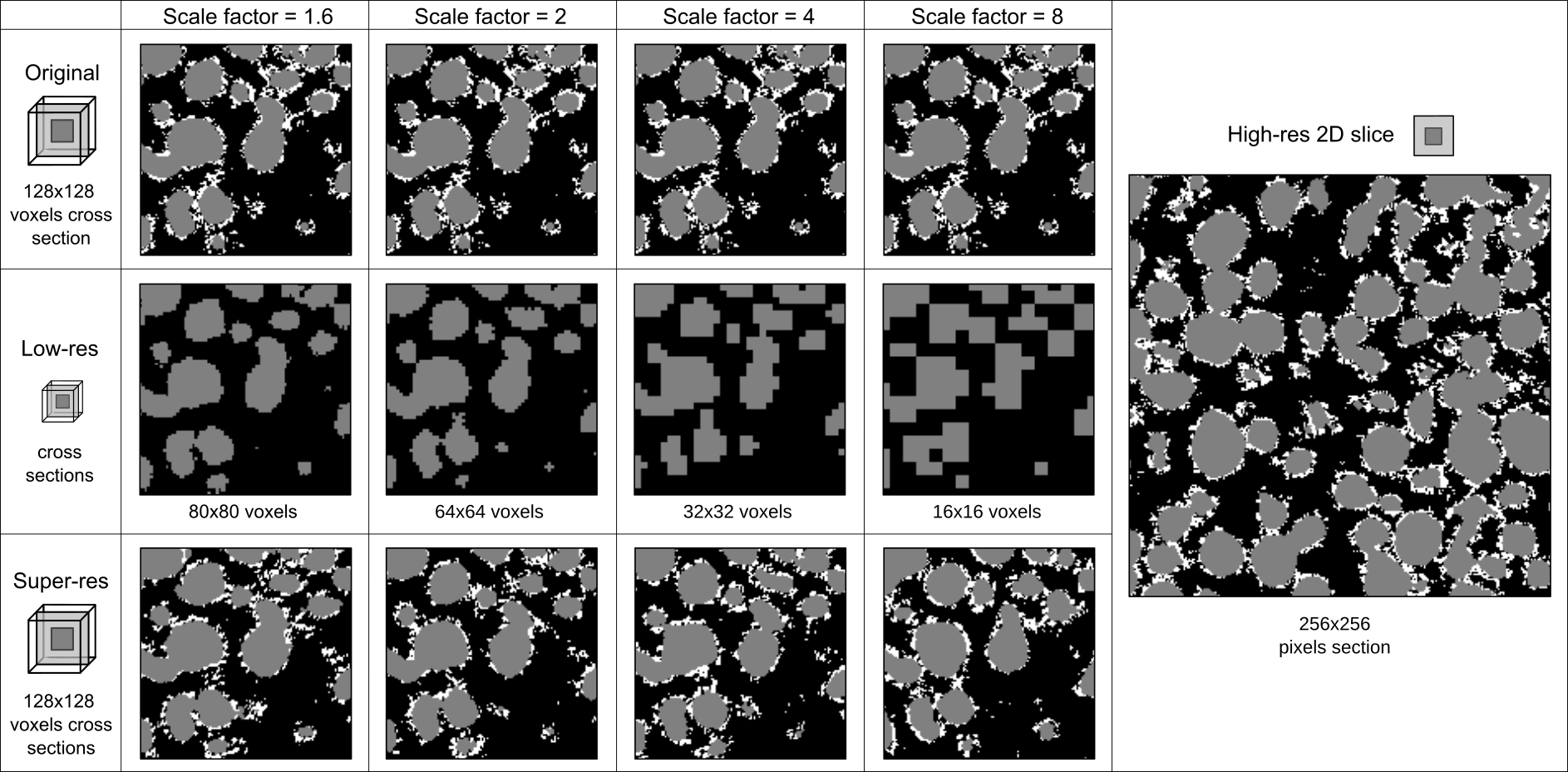}
    \caption{The results from exploring the effect of different scale factors. For each scale factor, the inputs for the model were a high-res 2D slice, whose center can be seen on the right, and a low-res volume which was blurred and down-sampled accordingly to the scale factor from the original high-res volume. Every column shows a different experiment for a different scale factor. The top row shows a cross section of the original volume, the middle row shows a cross section of a corresponding position in the low-res volume and the bottom row shows a cross section of a corresponding position in a generated super-res volume. }
    \label{fig:nmc_different_scale_factors}
\end{figure*}

The same {\em SliceGAN} model was also used to independently generate a 2D high-res three-phase slice containing $2048^2$ pixels. As explained in the introduction, the decision to merge the pore and binder phases in the low-res volume stems from the fact that they are very difficult to distinguish using XCT due to the very low X-ray attenuation of the binder compared to the active material.

Figure \ref{fig:Fig1-NMCimages} illustrates the workflow and effectiveness of the model. The different phases of the cathode shown are pore (black), active material (gray) and binder (white). In addition to demonstrating the generated super-res of the low-res volume, this scenario also exhibits the 3D in-painting of a phase present only in the high-res 2D slice. 

Figure \ref{fig:nmc_metrics} shows a quantitative comparison between the original high-res volume and the super-res volume. In the metrics comparison, the ultimate objective for the super-res volume is to be in good agreement with the 3D high-res ground truth volume, but since it only sees the 2D image as the high-res input, the agreement of the metrics depends on the representativeness of the 2D image to the ground truth volume. As can be seen, except for the pore/binder interphase surface area, the 2D high-res image is representative of the properties of the 3D volume, thus the super-res volume is in good agreement with the ground truth volume on all of these metrics. Interestingly, since the 2D high-res image is the only information for the high-res binder characteristics for the reconstruction, the super-res volume has good agreement with it for the Pore/Binder interphase surface area, slightly more than the 3D original volume. This good agreement with the 2D high-res image highlights the accuracy of the proposed method, as well as the importance of using a representative 2D slice for an accurate 3D reconstruction.

As described in more detail in Section \ref{Methods}, the model is able to support a variety of different scale-ratios between the resolutions of the two input images. Figure \ref{fig:nmc_different_scale_factors} shows the same NMC sample reconstructed from a range low-res inputs derived from different scale factors. It is clear that as the scale factor increases, the fidelity to the low-res volumes reduces, which is to be expected. However, all of the super-res results look like valid cathode regions, matching the high-res 2D slice characteristics.

For the extreme case of scale factor 8, 1 voxel in the low-res volume corresponds to $8^3=512$ voxels in the high-res volume. This naturally leads to a large loss of information when converting from the original volume to the low-res volume. As a result, there are a huge number of different reconstruction options for the super-res volume, all of which can be realistic and statistically similar to the original volume.

It is important to note in addition that Figure \ref{fig:nmc_different_scale_factors} only shows one of the many possible super-res outputs that could have been generated for each scale factor. Supplementary Information Figure \ref{fig:NMC_different_noise_seeds} shows the existing variability from different noise inputs for creating super-res volumes. This highlights the stochastic nature of the approach, as will be explained in the Methods section. The figure also shows a comparison to the original volume from which the low-res volume was created. Also, to quantify the change in resolution from the low-res to the generated super-res volumes, Supplementary Information Figure \ref{fig:fft_magnitude} shows a comparison in Fourier Space between the images using FFT magnitude maps.

\subsection{Case 2: Validation using an anisotropic, two-phase separator dataset}

\begin{figure*}[!htb]
    \centering
    \includegraphics[width=1\textwidth]{ 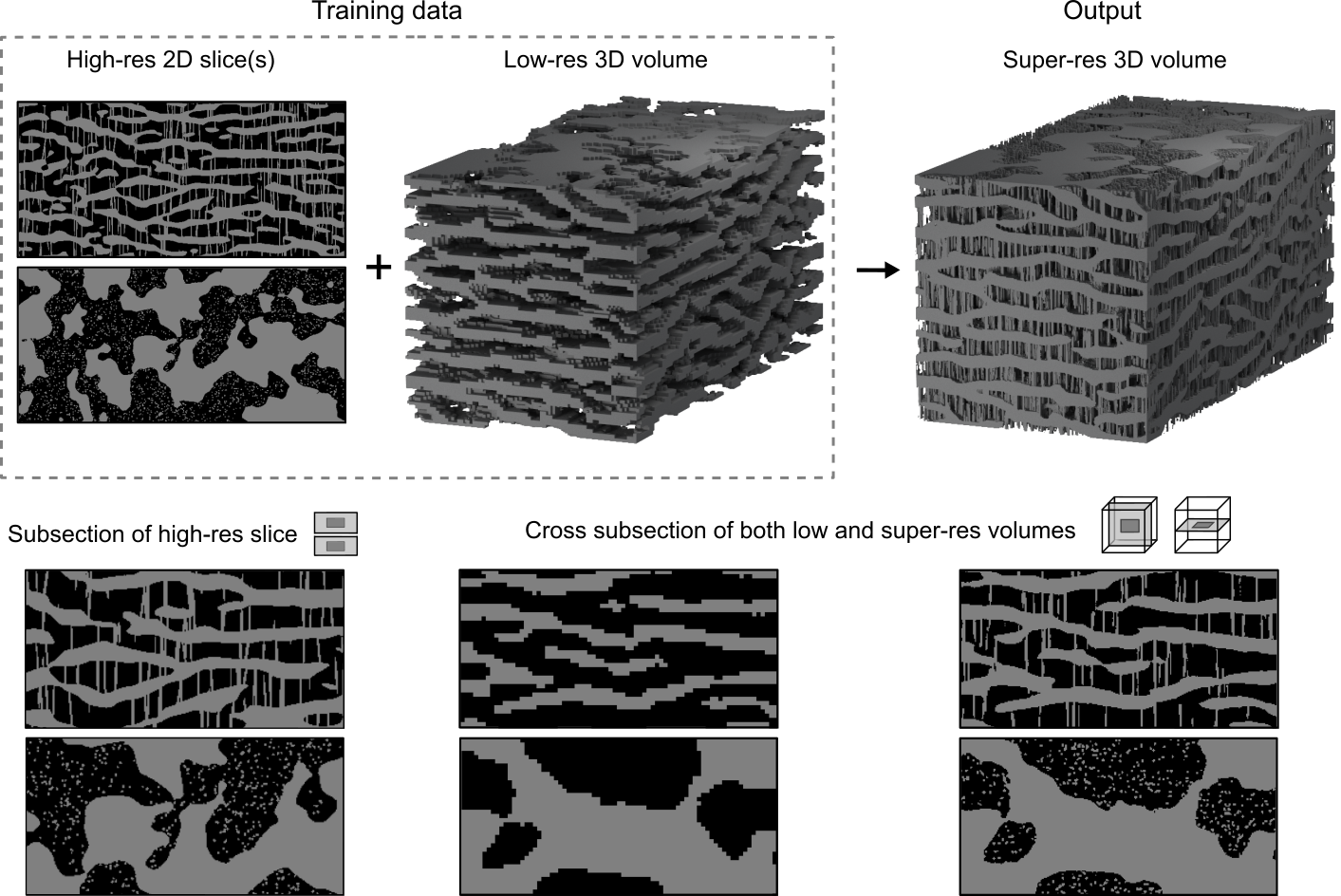}
    \caption{The model results for an anisotropic material, here producing a super-res 3D battery separator material \cite{finegan2016characterising} with a scale factor of 4. The different model inputs and outputs are the same as in Figure \ref{fig:Fig1-NMCimages}, with the additional input of high-res 2D slices from all three perpendicular directions. Since two perpendicular directions have the same properties (upper slice in the top left corner), for simplicity only two high-res 2D slices are shown out of the 6 facets used for training data.}
    \label{fig:separator_outline}
\end{figure*}

To present the flexibility and diverse capabilities of the method, an anisotropic material was also investigated. For this, an open-access dataset of a porous polymer battery separator material was chosen \cite{xu2020microstructure}. In the previous example, a new phase was introduced in the 2D high-res slice that was not present in the 3D low-res volume. However, an alternative scenario is that the same phase displays new features when imaged at higher resolution. The separator material consists of thick lamellae connected by thin fibrils.

Although the XCT data collected by Finegan \textit{et al.} \cite{finegan2016characterising} was able to capture the lamellae, it did not have sufficient resolution to observe the fibrils. However, Xu \textit{et al.} used a stochastic method to add fibrils \cite{xu2020microstructure} based on statistical information taken from an SEM image. In this experiment, the high-res 3D separator data is $624\times300\times300$ voxels of the lamellae and stochastically added fibrils. The low-res volume is a down-sampling of only the lamellae with a coarsening scale factor of 1/4. Together with the low-res volume, the input also consists of high-res 2D slices, which is the set of all 6 outer facets of the high-res 3D volume (analogous to capturing 2D images of the outer surfaces of a volume). These slices together sum to very limited amount of image training data (less than one megapixel) compared to the amount generally required for deep learning imaging applications \cite{wang2020deep}. The fact that the model still performs well is likely explained by the random sampling method used on the high-res input data, so a single pixel can appear in many training instances. However, using fewer facets did result in an unsuccessful super-res reconstruction, indicating the existence of a lower threshold for representativity and successful reconstruction in this scenario.

\begin{figure*}[!t]
    \centering
    \includegraphics[width=\textwidth]{ 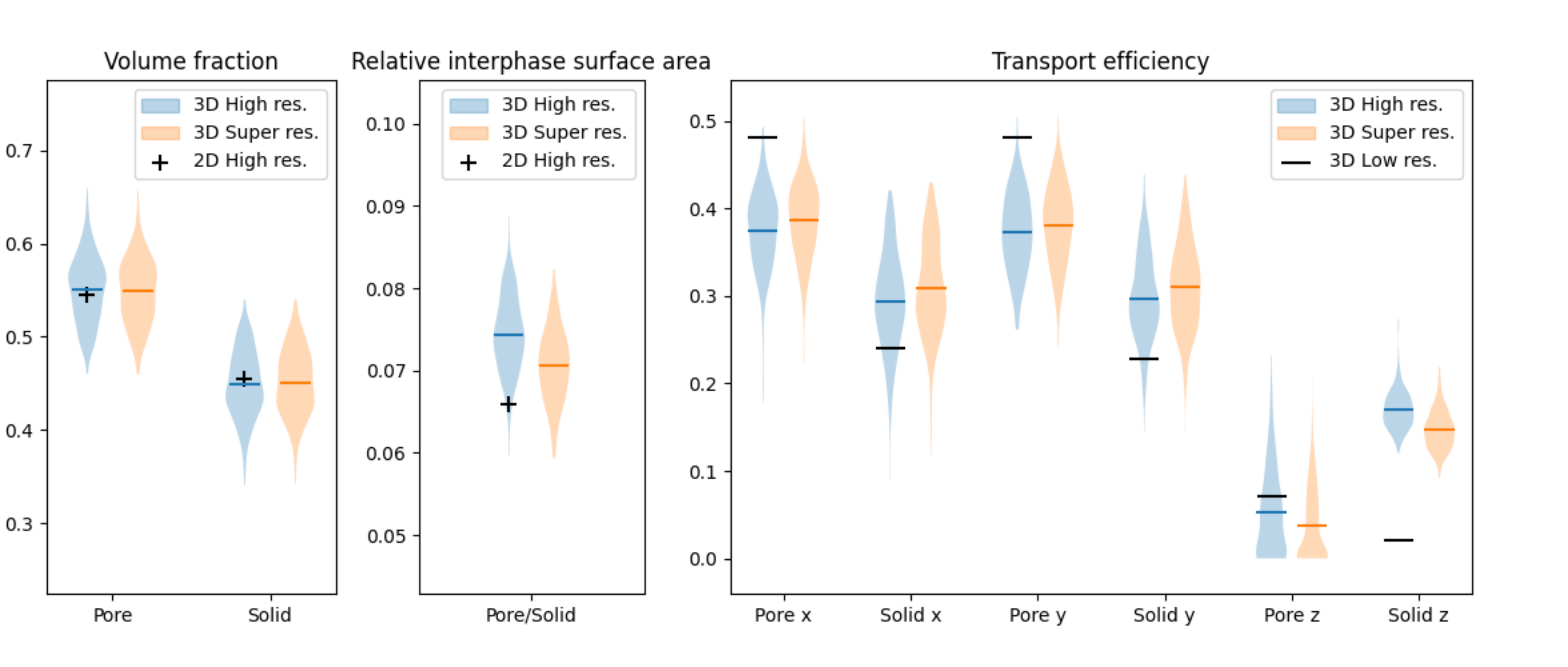}
    \caption{Statistical comparison of key mesostructural metrics between the original high-res 3D volume, the high-res 2D slices and the reconstructed super-res 3D volume for the anisotropic separator material in Figure \ref{fig:separator_outline}. The metrics description is the same as in Figure \ref{fig:nmc_metrics}. Since the material is anisotropic, transport efficiency measurements were taken along all different axis. The figure also shows the transport efficiency measurements of the low-res volume, and the added information  gained by super-resolving.}
    \label{fig:separator_metrics}
\end{figure*}

As can be seen in Figure \ref{fig:separator_outline} and the corresponding metrics comparisons in Figure \ref{fig:separator_metrics}, although the limited training high-res 2D data may have slightly harmed the agreement of the metrics, the facets of the original high-res volume had enough high-res information for the successful anisotropic super-res reconstruction. As in case study 1, the surface area of the high-res 2D slice was offset from the global 3D high-res value. The surface area of the 3D super-res lies between the high-res 2D value and the original volume data which the low-res data is derived from. This balance demonstrates the ability of the generated volume to accurately reproduce the statistics of the HR 2D slice while also resemble the low-res 3D volume training data.

\subsection{Case 3: Validation using data from a fuel cell anode with three-phase low-res data and super-sampling}

\begin{figure*}
    \centering
    \includegraphics[width=1\textwidth]{ 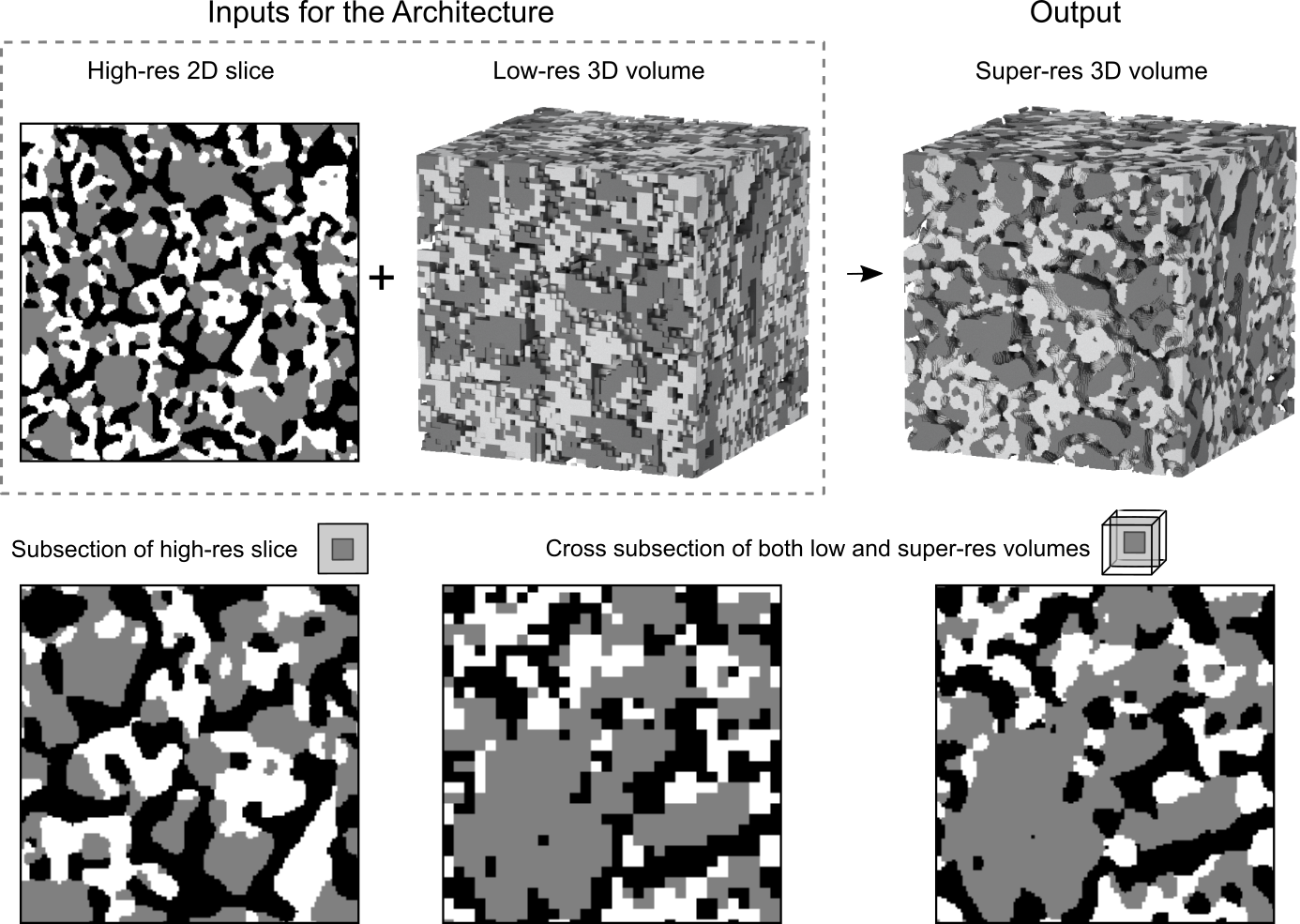}
    \caption{The SuperRes model inputs and output evaluated on the fuel cell anode FIB-SEM dataset (Case study 3) \cite{hsu2018mesoscale}, with a scale factor of 4. The different phases are pore (black), nickel (gray) and ceramic (white). In this example, the low-res is being super-sampled, rather than super-resolved. The arrangement of images is the same as in Figure \ref{fig:Fig1-NMCimages}.}
    \label{fig:SOFCimages}
\end{figure*}

To showcase several additional features of the SuperRes method, an open-access image dataset of a solid oxide fuel cell anode material was chosen \cite{hsu2018mesoscale}. The images were obtained through FIB-SEM serial sectioning and, as can be seen in Figure \ref{fig:SOFCimages}, this material has three phases: pore (black), metal (gray) and ceramic (white). In order to demonstrate super-sampling, instead of super-resolving, all three phases are present in the low-res image.

An additional dimensionless metric called the normalised Triple Phase Boundary (TBP) density was calculated for this material, as it is of particular relevance to the performance of SOFC anodes. Each voxel edge makes contact with four voxels and an edge where all three phases are present in its neighbours is considered to be a TPB. As with the other metrics, the calculation was performed using \textit{TauFactor} \cite{cooper2016taufactor} on 256 randomly sampled $64^3$ volumes. The mean and standard deviation for cubes from the training set were 0.01087 and 0.00121 respectively; whereas they were  0.01028 and 0.00126 respectively for the super-res generations. The exceptional similarity of these two distributions is further evidence for the effectiveness of the SuperRes approach.

It is important to note the distinction between under-sampled data (\textit{e.g.}, for SEM imaging, raster step size $>$ electron beam spot size) and under-resolved data (\textit{i.e.}, raster step size $<$ electron beam spot size) as this will impact how a ``low resolution” image should be interpreted by the loss function in the SuperRes algorithm. If the data is under-resolved, we apply a Gaussian blur of the appropriate size (before downsampling) in order to achieve the relevant degree of spatial mixing. However, if the data is under-sampled, the super-res voxels should instead be sampled at the appropriate spatial interval and compared directly to the low-res. This option is further explained in the Methods section.

\begin{figure*}
    \centering
    \includegraphics[width=1\textwidth]{ 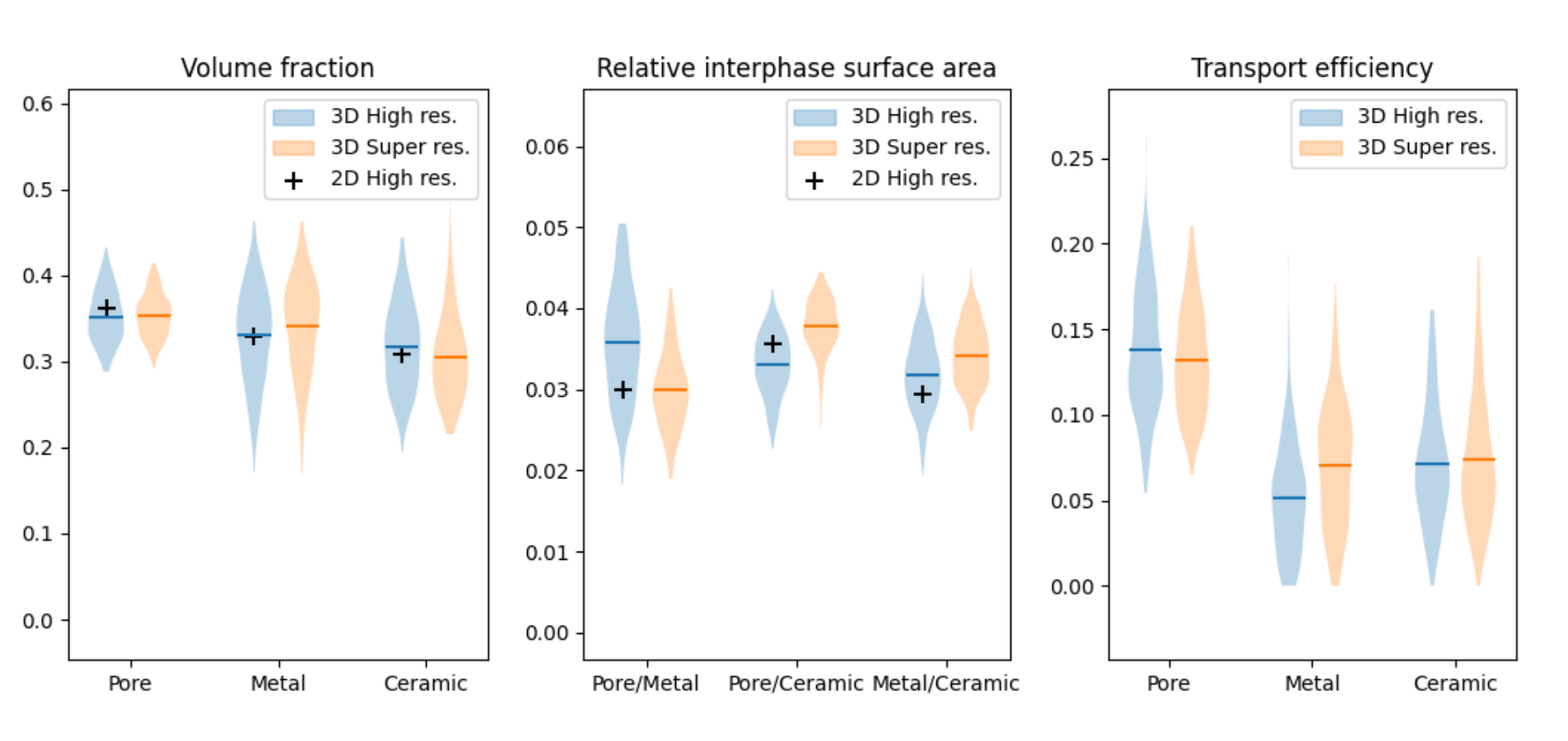}
    \caption{Statistical comparison of key mesostructural metrics between the original high-res 3D volume, the high-res 2D slices and the reconstructed super-res 3D volume for the SOFC anode material\cite{hsu2018mesoscale} in Figure \ref{fig:SOFCimages}. The metrics description is the same as in Figure \ref{fig:nmc_metrics}.}
    \label{fig:SOFCmetrics}
\end{figure*}

In the first two case studies, a Gaussian blur was applied before down-sampling, to simulate what might have been observed using lower resolution XCT imaging (i.e. a weighted volume average of regions of voxels). However, in Case 3 it is assumed that both the low-res and high-res data were collected using SEM. In an SEM, unlike XCT, the ``spot size" of the incident electron beam can be varied independently of the raster step size used to construct the image. So, each voxel in the low-res image represents the same amount of material as in the high-res, rather than an average of a larger region, like it would in XCT. This requires a simple modification to the way in which the data is downsampled when calculating the voxel-wise loss, excluding the Gaussian blur as described in the Methods section. Figures \ref{fig:SOFCmetrics} and \ref{fig:SI_2pc_cld} show a good match in measured statistics between the training data and original volume, although perhaps not quite as good as in the first two cases.

Supplementary Information Figure \ref{fig:SOFC_x8_different_noises} shows the existing variability from different noise inputs for creating super-res volumes with a high scale factor of 8.

\subsection{Case 4: Demonstration using a pair of complimentary datasets from a battery cathode}

\begin{figure*}
    \centering
    \includegraphics[width=1\textwidth]{ 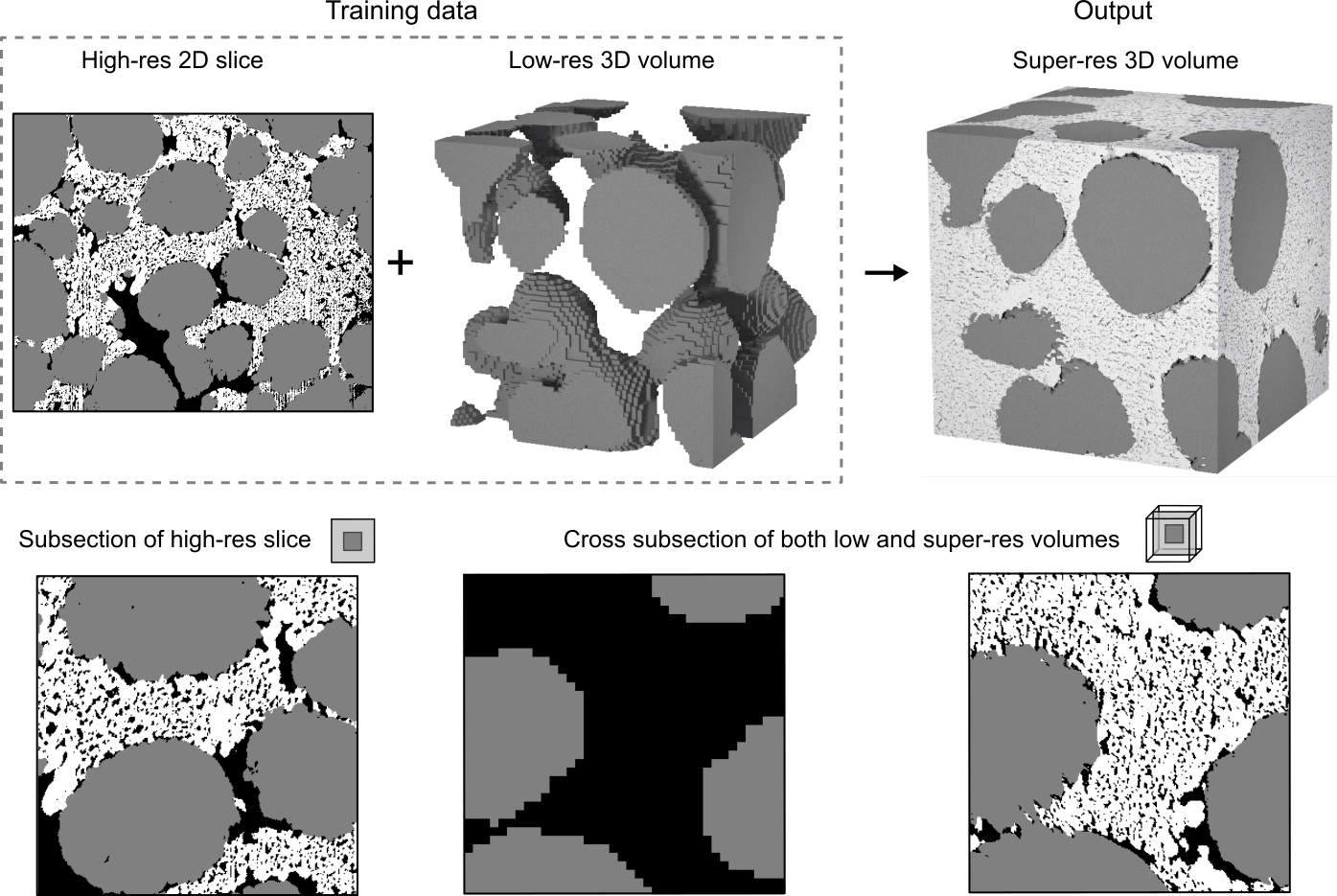}
    \caption{The model results for a demonstration isotropic material, combining XCT low-res volume and SEM high-res 2D slice, with a scale factor of 8. Resulting in a super-res volume with nanoscale resolution that is hard to impossible to achieve with current imaging techniques.}
    \label{fig:SEM_outline}
\end{figure*}

The previous case studies were designed to validate the accuracy of the method using experiments that had the ground truth high-res 3D volume from which the complementary training images were derived. This way the generated super-res volumes could be quantitatively compared to the ground truth. Once the performance has been validated, a natural next step is to test the method on a real-world problem, where the desired ground truth high-res 3D volume does not exist, but all of the necessary information is contained in a pair of complimentary datasets.

For the cathode dataset in the first study, the binder phase was added stochastically to the XCT imaged volume based on statistical information from an SEM image of the same material. In this case study, we will use the XCT dataset without the added binder phase as the low-res input, leaving only a map of the active material phase, as can be seen in middle top of Figure \ref{fig:SEM_outline}. 

The SEM image (secondary electrons) of the same cathode sample used to extract the binder statistics in \cite{usseglio2018resolving} will be used as the high-res 2D input. The high-res SEM image obtained from the same sample offered sufficient contrast to distinguish between the CBD phase from the pores, allowing it to be segmented into three phases. This task was performed using Weka segmentation software \cite{arganda2017trainable} accompanied by some manual finishes, to try and account for the ``pore-backs'' problem described in the introduction. The entire three-phase segmentation result can be seen in the upper left corner of Figure \ref{fig:SEM_outline}, while the original SEM image alongside the segmentation can be seen in the Supplementary Information Figure \ref{fig:sem_segmentation}, along with some further explanation of the segmentation approach in the figure's caption.

Here, other than a real representation of the CBD phase, another difference to the first case study is the resolution and the scale factors used. While in the first case study the low-res volume had a voxel lengthscale of 1.6 $\muup$m and the high-res 2D slice had a pixel lengthscale of 400 nm (scale factor of 4), in this case study the low-res volume has a voxel lengthscale of 400 nm and the high-res 2D slice has a pixel lengthscale of 50 nm (scale factor of 8). To the authors knowledge, this is the highest resolution Li-ion cathode mesostructure volume dataset that contains a realistic representation of all three phases currently available in the literature.

We mention that although the SEM image had sufficient information to observe the CBD's nanoscale features, it had many unwanted artifacts that damaged the true distribution of the phases in the image. As well as the ``pore-backs'' issue, which may cause the binder to appear over represented in the SEM image, an additional artefact referred to as 'curtaining' can occur when FIB is used to prepare cross-section for SEM imaging. Vertical streaks are formed by the ion beam milling through phases with different densities, which can make segmentation more challenging \cite{lemmens2011fib}. This was not totally accounted for during the manual correction of the segmentation and can be most clearly seen in the bottom right of the 2D input image in Figure \ref{fig:SEM_outline}. However, we find that this anisotropic artifact is hardly found in the output super-res volume. This positive finding can be linked to the isotropic rotational data augmentation we performed as a preprocessing step, which will be described in the next section.

The limited high-res 2D training data in this case is not representative of the low-res 3D volume, which is noticeable from the high active material volume fraction of 0.58 compared to 0.40 in the low-res volume. This discrepancy could have originated either from the small spatial area of the high-res 2D image or the under-classification of active material in the XCT segmentation \cite{usseglio2018resolving}. We believe that this unrepresentative slice led to slight mismatch in the comparison of the pore structure between the generated volume and high-res slice. As can be seen in the high-res slice, pore tends to be present in the vicinity of particles, especially between close adjacent particles. The low active material volume fraction and and the low prevalence of close adjacent particles in the low-res volume led to lower pore volume fraction in the generated volume.

However,  it is clearly visible that the binder phase in the super-res volume qualitatively matches the elaborate structure of the binder phase in the SEM image shown in Figure 6. Furthermore, the contrast to the geometry of the binder phase that was added statistically (based on metrics from the same SEM image) in the open-access data as shown in Figure 1 is stark. Although there might be inaccuracies comparing statistics between different resolution samples, our alternative method of stochastically adding binder to the XCT volume have binder volume fraction of 0.506 compared to 0.111 in \cite{usseglio2018resolving}, and the binder phase percolates with a transport efficiency of 0.242 compared to the almost non-percolating 0.008 transport efficiency of the binder phase in \cite{usseglio2018resolving}.

To further test the statistical alignment of the generated volumes to the original volumes, three more functional descriptors \cite{stoyan2013stochastic} were performed: intraphase and interphase 2-point correlation functions and chord length distribution. These metrics have all shown good agreement between the original and generated volumes for the first three case studies, and can be viewed in Supplementary Information Figure \ref{fig:SI_2pc_cld}.

\section{Methods}
\label{Methods}

The SuperRes model was built on the development of {\em SliceGAN}\cite{kench2021generating}, a generative adversarial network architecture that generates 3D mesostructure from a 2D slice. In SliceGAN, the input of the Generator ($G$) is random noise and the output is a one-hot encoded 3D mesostructure which is then sliced orthogonal to $x$, $y$ and $z$ axes to feed the Discriminator ($D$) with 2D images. Here, as well as feeding $G$ with random noise, the input of $G$ also contains a low-res 3D mesostructure which we wish to super-resolve (or possibly super-sample) and augment. The architecture of $G$ is changed accordingly, whilst $D$ remains essentially unchanged. Although it is a reasonable assumption that $G$ would learn to use the low-res input structure to generate a high-res reconstruction with similar features, this is not guaranteed to be the case. This is because the standard GAN loss function does not include a measure of the similarity between the input and output image. It is thus possible for the generated 3D images to look very different from the low-res image. 

To constrain $G$ to generate images that will have the same features as the low-res input image, we add a loss term given by the voxel-wise mean squared error (MSE) between the low-res image and a down-sampled version of the super-res image. If the imaging technique used to obtain the low-res images is truly ``under resolving'' the high-res features, then it is reasonable to apply a Gaussian blur before the down-sampling, such that the low-res can be compared against a weighted volume average of the voxels that correspond to each voxel in the low-res.

The standard deviation and size of the 3D Gaussian kernel used to blur the super-res generations can be tuned to more closely represent the imaging conditions used. However, the default mode, as applied in Cases 1 and 2, used a 3D kernel with odd kernel size $k$ defined by the following equation:

\begin{equation*}\label{running_time_tsitsas}
    k=\begin{cases}
    \lceil sf \rceil & \text{if } \lceil sf \rceil \text{ is odd} \\
    \lceil sf \rceil -1 & \text{if } \lceil sf \rceil \text{ is even}
    \end{cases}
\end{equation*}

Where $sf$ is the scale factor and the Gaussian kernel weights are distributed by the default PyTorch \cite{paszke2019pytorch} $\sigma=0.3\cdot((k-1)\cdot0.5-1)+0.8$ for the appropriate kernel size $k$. The super-sampling approach explored in Case 3 can be thought of as a Gaussian blur with a standard deviation of zero (although, in reality, it makes more sense to just turn the blur off). For completeness, it should be noted that a Gaussian blur with very large (or infinite) $\sigma$ would produce a kernel with uniform weighting, which is analogous to pixel ``binning''. However, all of these case produce very similar results for scale factors $\leq 4$.

Since the low-res input to $G$ is a segmented volume, after down-sampling the generated volume a low temperature softmax function was applied instead of a threshold (step) function to result in a \textit{near} one-hot encoded phase map, which keeps the voxel-wise loss differentiable. Other than the addition of the voxel-wise loss, the Wasserstein loss function \cite{arjovsky2017wasserstein} remains similar to SliceGAN. This procedure is outlined in Algorithm \ref{alg:SuperResAlgo} and in the Supplementary Information Figure \ref{fig:training_outline}.

\begin{algorithm*}[htb!] 
	\begin{algorithmic}[1]
	\Require $G$, the generator function; $D$, the discriminator function; $LR$, the low-res volume; $HR$, the high-res 2D slice; $sf$, the scale factor between $LR$ and $HR$; $b$, threshold for the voxel-wise loss  (default $0.005$); $c$, the coefficient of the voxel-wise loss multiplication (default $10$); $\sigma$ the downsampling function: a Gaussian blur followed by a trilinear downsample with a scale factor of $sf$, followed by a differentiable near one-hot encoding (a softmax with low temperature); All batch operations and optimization and gradient penalty parameters are not shown for simplicity, we refer the reader to the codes at the project's repository for specific parameter details, including details about the implementation of the gradient penalty.
	\Preprocessing $HR\leftarrow$ Data augmentation step, expanding $HR$ into all 8 different possible mirrors and rotations of $HR$ by $\ang{90}$ multiples.
	\Statex
		\While {\text{weights of $G$ did not converge}} 
		\State $lr\leftarrow$ Sample a $(\frac{64}{sf})^3$ voxels cube uniformly from $LR$.
		\State $z\leftarrow$ Sample a $(\frac{64}{sf})^3$ voxels cube of noise from normal distribution with mean $0$ and standard deviation $1$.
		\State $lr\leftarrow concat(lr,z)$ Concatenate the low-res and noise cubes along the phase dimension.
		\State $sr \leftarrow G(lr)$ Generate a $64^3$ super-res volume.
		\State $sr_{cropped} \leftarrow$ Crop $4$ facets from every outer plane of $sr$ into a $56^3$ volume.
		\State $sr_{slices}\leftarrow slice(sr_{cropped})$ Slice the volume $sr_{cropped}$ in all perpendicular planes into $56\cdot3$ 2D slices. 
		\If {Training of $G$ (Only every 5 iterations)}
		\State $l_{vw}\leftarrow MSE(lr,\sigma(sr))$ Voxel-wise loss between low-res and downsample of Gaussian blurred super-res volumes.
		\If {$l_{vw}<b$}
		\State $l_{G}\leftarrow-D(sr_{slices})$
		\Else
		\State $l_{G}\leftarrow-D(sr_{slices})+c\cdot l_{vw}$
		\EndIf
		\State Backpropagate and update the weights of $G$ from the loss $l_{G}$.
		\Else {~Training of $D$~}
	    \State $hr\leftarrow$ Sample a $56^2$ pixels square uniformly from $HR$.
	    \State $l_{gp}\leftarrow$ GP($sr_{slices}$, $hr$, $D$) Calculate the gradient penalty regularization based on real and fake outputs. 
	    \State $l_{D}\leftarrow D(sr_{slices})-D(hr)+l_{gp}$
	    \State Backpropagate and update the weights of $D$ from the loss $l_{D}$.
		\EndIf
		\EndWhile
		\Return The super-res volume of the full low-res volume $\mathbb{G}(LR)$.
		\caption{{\footnotesize SuperRes algorithm for isotropic materials}}
		\label{alg:SuperResAlgo}
	\end{algorithmic}
\end{algorithm*}

We emphasize that both the low-res and high-res input data for the model are one-hot encoded (i.e. already segmented into channels, rather than the raw grayscale). However, the model could be modified in future to utilise grayscale or even colour image data directly if some advantage was identified.

Any overfitting of $D$ on the high-res training data can disrupt the learning process of $G$. This failure mode will enforce $G$ to generate volumes with high voxel-wise loss to the input volumes, in an attempt to copy the training data.

Data augmentation can increase the apparent diversity of the training set and therefore improves the quality of the trained model by reducing overfitting. This is particularly important when training data is hard to acquire and scarce. For example, if the material is assumed to be isotropic then a single 2D slice can be transformed into 8 slices of the same size, constituting of all the different mirrors and $\ang{90}$ rotations of the original image. It should also be noted that the 2D high-res training data does not need to come from a single large image, but can be from a collection of smaller images, which may be much easier to obtain. 

From our experiments, lack of volumetric low-res data was never a barrier for successful training thus no data augmentation of the low-res 3D input was performed but can easily be implemented. This can be explained by the extra dimension in 3D data that multiplies the number of different samples, or the importance of diversity in the fine features of the 2D high-res input compared to the coarse, often repeated, 3D low-res structure input.

The voxel-wise loss between the low-res and super-res volumes is only used to update the parameters if the voxel-wise loss is above the 0.5\% threshold. This threshold is introduced to provide flexibility for the super-res output to distance from the blocky corners of the phases present in the low-res volume. More flexibility comes from the concatenated input noise layer to $G$ and is necessary for a successful reconstruction, as shown in the Supplementary Information Figure \ref{fig:importance_of_noise}.

Algorithm \ref{alg:SuperResAlgo} is for isotropic materials and so a minor modification is required to generate anisotropic materials. This modification is identical to that used in SliceGAN \cite{kench2021generating}, where $3$ different discriminators can be used for the x-y, y-z and x-z directions. Note that the last step of the algorithm output is possible due to the convolutional design of the generator that is detailed in Supplementary Information Figure \ref{fig:g_structure}, that allows for any size of input volume.  

For clarification, in each different case study experiment a ‘new’ model is trained independently without any prior knowledge (i.e. from a random initialisation). It is the same architecture but with different weights after training, optimised for the specific training data. A future direction for improving the method could be to use transfer-learning from a model trained on some general mesostructural data, shortening the learning and convergence time of the specific application.

Due to the adversarial nature of GANs, training can be unstable. For example, overfitting of the discriminator on the high-res 2D training data 
can in turn degrade the performance of the generator, making its outputs worse from the perspective of mesostructural generation. Furthermore, Wasserstein GANs do not have trivially definable convergence criteria \cite{mescheder2018training}. Ultimately, since the measure of whether the model is performing well depends on what you wish to do with the outputs, both the instability and arbitrary convergence can be addressed by extracting cheap metrics of interest during the training process (e.g. volume fractions and surface areas) and using these to help determine when to stop training.

Around the edges of the super-res volume the uncertainty high. This is because the low-res volume does not contain any information about the regions adjacent to its boundaries. So, for example, in the case of the cathode, a pore just inside the boundary may in reality be next to a particle of active material that was just outside the imaged region. This would effect the morphology of the binder phase nearby. This concept is reflected in the non-uniform information density near the edges of the volume, as described in \cite{kench2021generating}. As such, it is advisable for users of the SuperRes method to crop their volume by the equivalent of 1 voxel of the input volume from the generated volume (e.g. in 4$\times$ resolution increase, remove 4 layers from each face), as advised in the github repository manual of this project. For this reason, the generated volumes edges are being cropped before being discriminated.

The model is able to support a variety of different scale-ratios between the resolutions of the two input images. Specifically, the model supports all rational scale factors in the range between 1 to 8 that divide 64 without a remainder, which are all scale factors $\frac{64}{d}$ for any integer $d$ in the range $8\leq d\leq 64$. This set of scale factors can be expanded and is only limited by current implementation.

Our proposed model does not implement the classic style-transfer algorithm by Gatys \textit{et al.} \cite{gatys2016image}, in part because of the 2D to 3D dimensionality mismatch of the style representation. However, our approach is not too dissimilar to the framework in \cite{gatys2016image}, as the loss of $G$ consists of a linear combination of the content representation (the voxel-wise loss) and the style representation ($D$'s score for fake images).

\section{Discussion}

The results of the four case studies make a strong case for the utility and potential impact of this GAN based super-resolution method. The first, second and third case studies demonstrated excellent agreement between the metrics calculated on the 3D high-res ground truth and super-res outputs. The excellent agreement of the transport efficiency metric is particularly notable as it is an inherently 3D property. This highlights the power of this method to perform both dimensionality expansion as well as data fusion, as this property could not be accurately calculated on the low-res 3D volume or calculated at all on the high-res 2D data. 

There is a small discrepancy between some of the 3D high-res and 3D super-res metrics, with the super-res metrics more closely aligned to the 2D high-res metrics rather than 3D high-res metrics. Moreover, certain metrics are only contained in the high-res 2D training data, such as the binder phase fraction in the NMC sample. These results illustrate how the model is learning the distribution of the 2D high-res training data whilst being constrained through the super-resolution process to match the gross arrangement enforced by the low-res input.



\begin{figure}[t]
    \centering
    \includegraphics[width=1\textwidth]{ 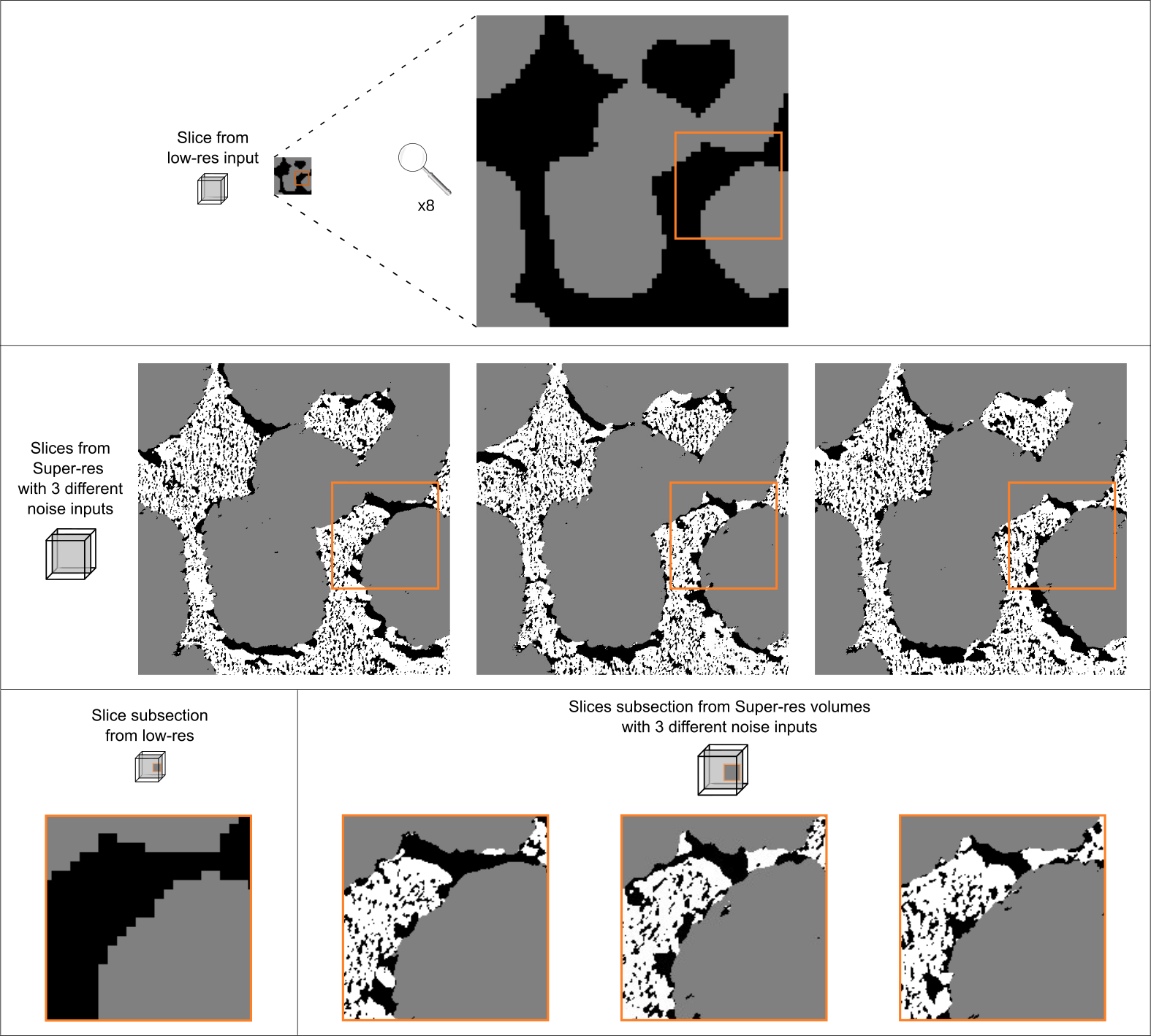}
    \caption{Exploring the effects of different noise inputs for the generation of different super-res volumes, generated by the model of Case study 4. The top row shows a slice of the low-res volume and a $\times8$ stretched magnification for easier comparison. This low-res volume input acts as an input for the model, together with random noise. The middle row shows a corresponding slice (\#2 out of possible 8) of the low-res volume in 3 different super-res volumes, generated with 3 different random noise inputs. The bottom row shows the same subsection of all the slices, showing the fine feature variability that exists as a result of different noise inputs.}
    \label{fig:SEM_different_noise_seeds}
\end{figure}

\subsection{Stochasticity}

It is interesting to note that it is possible to implement this super-res method using just a low-res volume as the input to $G$, without concatenating a layer of noise. In this scenario, the model is deterministic and if it is well trained, its output should represent a likely underlying mesostructure associated with each neighbourhood of the input. Yet, the addition of noise offers three advantages over the deterministic approach: 
\begin{itemize}
    \item Limiting $G$ to a single output for a given low-res input could lead to unwanted homogeneity of the resulting mesostructure, especially when the low-res input contains many repeated patterns. This is exemplified for an input featuring only one phase, as shown in Supplementary Information Figure \ref{fig:importance_of_noise}. Unlike the heterogeneity of features generated with the stochastic input, without added noise $G$ is only able to output a single pattern across the pore volume, resulting in a failure mode. The figure also shows the scenario with large patches of pore between particles and a similar but more complex failure mode is observed. 
    
    \item There are two ways in which overfitting of $D$ can occur, either $D$ learns the true dataset as described in the Architecture section, or $D$ can learn the output distribution of $G$. In this scenario, after $D$ learns $G$'s outputs, it can simply classify this dataset as fake, and all other data as true. By adding noise, $G$ can learn to use it to generate greater diversity in its generated outputs to overcome this kind of overfitting. This is especially relevant when only a small amount of low-res volume data is available. There are many possible super-res reconstructions for a single low-res volume, enlarging $G$'s output space thus helps avoiding overfitting of $D$ on $G$'s outputs.
    
    \item By varying the noise one can generate different mesostructures and explore the distribution of potential features and properties, allowing the user to quantify the uncertainty and richness of the super-resolution process. Additionally, the fact that the noise is spatially distributed with the input volume, allows for the creation of periodic boundaries of the properties that are governed by the noise input \cite{Gayon-Lombardo2020}. Therefore, latent space exploration of the noise input and its use in mesostructure optimization is promising as a subject of future research.
\end{itemize}

It is possible to try and replicate the capability of the proposed method using only the SliceGAN presented by Kench and Cooper \cite{kench2021generating}. This can be attempted by searching the latent space of a pretrained generator for a specific high-res 3D mesostructure whose down-sampled version and the input 3D volume have a low voxel-wise loss. However, this process is often unsuccessful, landing in unrealistic local minima far from the desired mesostructure. This is likely because although the latent space of a typical SliceGAN generator contains a very large space of possible mesostructures, it does not contain every possible arrangement of features. Thus, it is highly likely that a specific 3D image above a certain size is not contained within $G$. SliceGAN is only able to reconstruct mesostructures with the same statistical properties, as opposed to the exact high-res volume associated with the available low-res dataset. 

Any super-resolution approach will contain a degree of uncertainty in the generated result. One advantage of the SuperRes approach is that the expected variation in the generated volume can be explored by varying the noise input as can be seen in Figure \ref{fig:SEM_different_noise_seeds} and subsequent figures in the Supplementary Information. 

\subsection{Representativity}

During training, $G$ generates $64^{3}$ voxel cubes in such a way that, after convergence, every 2D perpendicular $64^{2}$ slice of the cube is indistinguishable by the discriminator from the real training data $64^2$ samples. After training, the final output is then generated using an arbitrarily larger input to the generator. For example, a low-res input volume of $128^{3}$ with a scale factor of $4$ will output a super-res volume size of $512^{3}$, while during training generating $64^{3}$ volumes required only $16^{3}$ low-res inputs. A possible limitation of the SuperRes method is that the large volume will fail to capture very long-range interactions, since it was only trained and measured by generating small volumes. 

The impacts of limited high-res training data are highlighted in the second and fourth case studies. We have not encountered low-res training data limits, nor do we expect this to be encountered by users of the model. This is because the low-res data typically captures a large FoV in 3D, which should provide plenty of data, especially after data augmentation.

At first glance, the fact that the low-res volume is used for both training and evaluating the model (by being used in generating the result), might seem problematic. However, in the context of super-resolution, generalization is not an issue because the model’s mission is to generate a super-res image of the low-res input data given. This also simplifies the requirements on the user as they are only required to submit two data-sets: The high-res 2D slice and the low-res 3D volume to be super-resolved; there is no need for a train/validation/test data division.

We have not rigorously defined what representative is in respect to the high-res training data. Representativeness is an illusive concept in the literature, partly because its definition is context dependent. For example, a sufficiently large volume for confidently establishing volume fractions may not be large enough to act as a representative domain for simulating electrochemical processes. Broadly, a representative volume of a macrohomogeneous material can be seen as one where the statistics of the volume are acceptably close to the global statistics of the material. However, this can be hard to establish given limited data. An investigation of the definition of representativity in energy materials will be presented in a future study.

In our high-res data, the isotropy is assumed based on our knowledge of the material, but not quantified. It could be that some degree of anisotropy was present. An implication for our work is that, for example, the binder phase in case study 1 might appear to be isotropic based on the three orthogonal slicing directions, but sampling a slice of a generated volume in an arbitrary direction may yield different statistics. However, we suspect this will only have a fairly modest impact on the resulting structures of the cases we explore, related to the level of anisotropy of the high-res input data.

The transport efficiency \cite{cooper2016taufactor} (the ratio of the phase fraction to its tortuosity factor) can be used to explore the longer range interactions in a large volume, because it is calculated by simulating steady-state diffusion through the entire volume between two parallel faces. A comparison was made between the transport efficiencies across the different phases of the super-res large volume and the mean of the small $64^3$ cubes in both the validation case studies, and no apparent difference was found, with $<0.01$ mean absolute difference. This positive finding can be linked to the nature of the super-resolution process, as the bulk structure of the material is already present in the low-res volume, retaining the long-range structure of the material. For example, the active material arrangement in the first case study, and the lamellae arrangement in the second case study.

\section{Conclusion}

The results presented in this paper show the ability of the SuperRes algorithm to faithfully synthesize a high-res, multi-phase and large FoV volume by fusing complementary 2D and 3D image data. This was shown across a range of lengthscales and phase sensitivities. Although this study focused on energy storage materials, the application of SuperRes is not limited to this domain, and could be extended to other materials and imaging techniques. For example, catalyst beds, rock formations, or snow and soil packing, as well as medical imaging.

By the complementary fusion of 2D and 3D image data, this method can overcome the physical and experimental constraints of contemporary 3D imaging such as contrast, resolution and FoV. With realistic, super-resolved, style-transferred 3D data, more accurate simulations and characterisations of materials can be performed. Finally, even when possible, imaging high-res 3D volumes can be expensive and time-consuming. SuperRes offers an alternative, exchanging expensive machinery and complex techniques with open-source software.

Future developments of this method might include adjusting the architecture such that it can more easily capture longer range relationships as the network can be deeper and the information density could be shared among a larger region. Other future work might include a combination with other common GAN approaches such as conditional GANs and transfer learning. The former will enable interpolation between mesostructures with differing properties, while the latter aims to speed up the training process. In this study, only imaging techniques that capture the geometry and phase have been used. An extension of this method could include a data fusion of techniques that capture a wider range of crystallographic, chemical and mechanical information, such as electron backscatter diffraction and energy-dispersive X-ray spectroscopy. Additionally, this model could be used in an algorithm that applies to an in-painting problem, whereby missing or erroneous data can be in-painted, removing the need to repeat long imaging processes containing small errors and avoiding throwing away potentially useful datasets. Thus, there is broad potential for expansion of this approach as a tool for material characterization and optimization.

\section*{Data Availability}

The study used open-access training data available from the following sources as described in the Supplementary Information Table \ref{tab:Scenarios}. All the generated volumes and their corresponding input data are available at \url{https://zenodo.org/record/7104555#.YyxPedJByGY}.

\section*{Code Availability}
The codes used in this manuscript are available at \url{https://github.com/tldr-group/SuperRes} with an MIT license agreement.

\section*{Acknowledgements}

This work was supported by funding from the EPSRC Faraday Institution Multi-Scale Modelling project (https://faraday.ac.uk/; EP/S003053/1, grant number FIRG003 received by AD, SK and SJC), and funding from the President’s PhD
Scholarships received by AD. Funding from the Henry Royce Institute's Materials 4.0 initiative was also received by AD and SJC to support the completion and extension of this study (EP/W032279/1). The Titan Xp GPU used for this research was kindly donated by the NVIDIA Corporation through their GPU Grant program to SJC. The authors would also like to thank Liam Yasin, Dominic Williamson, Owen O'Connor, Katie Zeng and Boyd Siripornpitak for their useful comments, as well as the rest of the tldr group.

\section*{Authorship}
AD designed and developed the code for SuperRes, trained the models, performed the statistical analysis and drafted the manuscript. SK, IS and SJC contributed to the development of the concepts presented in all sections of this work, helped with data interpretation and made substantial revisions and edits to all sections of the draft manuscript.

\section*{Competing interests}
The authors declare no competing interests.

\section*{References}
\addcontentsline{toc}{section}{References}

\def\addvspace#1{}

	\renewcommand{\refname}{ \vspace{-\baselineskip}\vspace{-1.1mm} }
	\bibliographystyle{abbrv}
    \bibliography{main}

\end{multicols}

\include{Sup_Info}

\end{document}

%% file: preamble.tex
\usepackage{geometry}
\usepackage{titling}
\usepackage[utf8]{inputenc}
\DeclareUnicodeCharacter{2212}{-}
\usepackage{amsmath}
\usepackage{microtype}
\usepackage{siunitx}
\usepackage{lineno}
\usepackage{authblk}
\usepackage{lscape}
\usepackage{graphicx}
\usepackage{subcaption}
\usepackage[font={footnotesize},labelfont=bf]{caption}
\usepackage[super,numbers,sort&compress]{natbib}
\usepackage{booktabs}
\usepackage{blindtext}
\usepackage{booktabs,multirow,array}
\usepackage{tikz}
\usepackage{verbatim}
\usepackage{chemfig}
\usepackage{gensymb}
\usepackage{textcomp}
\usepackage{setspace}
\usepackage{multicol}
\usepackage{pgfplots,pgfplotstable}
\pgfplotsset{compat=1.17}
\usepackage{afterpage}
\usepackage{xtab,afterpage}
\usepackage{amssymb}
\usepackage{rotating}
\usepackage{scrextend}
\usepackage{tcolorbox}
\usepackage{xr}
\usepackage[hidelinks]{hyperref}
\usepackage{cleveref}
\usepackage{algorithm}
\usepackage[noend]{algpseudocode}
\usepackage{enumitem}
\setlist[itemize]{leftmargin=*}
\usepackage{microtype}
\usepackage{float}

\algnewcommand\algorithmicinput{\textbf{Preprocessing:}}
\algnewcommand\Preprocessing{\item[\algorithmicinput]}

\usepackage{arxiv}

\usepackage{txfonts}
\usepackage{soul}

%% file: Sup_Info.tex


	






\section*{Supplementary Information}

\newpage

\begin{table}[!htb]

\resizebox{\textwidth}{!}{ 
\begin{tabular}{|cccccccccc|}
\hline
Case & Name & Isotropy & Data & Size (voxels number) & Shown in figure & Phases visible & Resolution & Field of view & Source \\ \hline
\multirow{6}{*}{1} & \multirow{6}{*}{\begin{tabular}[c]{@{}c@{}}Battery \\ cathode\end{tabular}} & \multirow{6}{*}{Isotropic} & High-res input & $2048^2$ & $256^2$ (Both Fig. \ref{fig:Fig1-NMCimages} and \ref{fig:nmc_different_scale_factors})& 3 & Medium & Large & \multirow{6}{*}{\begin{tabular}[c]{@{}c@{}}Both inputs were generated \\ using \cite{kench2021generating} from dataset \cite{usseglio2018resolving}.\end{tabular}} 
\\
 &  &  & Low-res input 1.6 & $320^3$ & $128^2$ (Fig. \ref{fig:nmc_different_scale_factors}) & 2 & low & Large &  
\\
 &  &  & Low-res input 2 & $256^3$ & $128^2$ (Fig. \ref{fig:nmc_different_scale_factors}) & 2 & Low & Large &  
\\

 &  &  & Low-res input 4 & $128^3$ & $60^3$ (Fig. \ref{fig:Fig1-NMCimages}) $128^2$ (Fig. \ref{fig:nmc_different_scale_factors})& 2 & V. low & Large &  \\
  &  &  & Low-res input 8 & $64^3$ & $128^2$ (Figure \ref{fig:nmc_different_scale_factors}) & 2 & V. low & Large &  
\\
 &  &  & Super-res output & $512^3$ & $240^3$ (Fig. \ref{fig:Fig1-NMCimages}) $128^2$ (Fig. \ref{fig:nmc_different_scale_factors})& 3 & Medium & Large &  \\ \hline
\multirow{3}{*}{2} & \multirow{3}{*}{\begin{tabular}[c]{@{}c@{}}Battery\\ separator\end{tabular}} & \multirow{3}{*}{Anisotropic} & High-res input & $624\times300\ (\times4)+300^2\ (\times2)$ & $624\times300\ (\times2)$ & 2 & High & Small & \multirow{3}{*}{\begin{tabular}[c]{@{}c@{}}High-res input \cite{xu2020microstructure} \\ Low-res input \cite{finegan2016characterising}. \end{tabular}} \\
 &  &  & Low-res input & $156\times75\times75$ & $156\times75\times75$ & 2 & Low & Large &  \\
 &  &  & Super-res output & $624\times300\times300$ & $624\times300\times300$ & 2 & High & Large &  \\ \hline
\multirow{3}{*}{3} & \multirow{3}{*}{\begin{tabular}[c]{@{}c@{}}Fuel cell\\ anode\end{tabular}} & \multirow{3}{*}{Isotropic} & High-res input & $1024^2$ & $256^2$ & 3 & High & Medium & \multirow{3}{*}{\begin{tabular}[c]{@{}c@{}}Both inputs were \\ obtained from dataset \cite{hsu2018mesoscale}.\end{tabular}} \\
 &  &  & Low-res input 4 & $64^3$ & $64^3$ & 3 & Low & Large & \\
 &  &  & All super-res outputs & $256^3$ & $256^3$ & 3 & High & Large &  \\
 \hline
\multirow{3}{*}{4} & \multirow{3}{*}{\begin{tabular}[c]{@{}c@{}}Battery\\ cathode\end{tabular}} & \multirow{3}{*}{Isotropic} & High-res input & $758\times628$ & $758\times628$ & 3 & High & Small & \multirow{3}{*}{\begin{tabular}[c]{@{}c@{}}High-res input (SEM)\\ and low-res input (XCT) \\ from dataset \cite{usseglio2018resolving}.\end{tabular}} \\
 &  &  & Low-res input & $240^3$ & $64^3$ & 2 & Medium & Large &  \\
 &  &  & Super-res output & $1920^3$ & $512^3$ & 3 & High & Large &  \\ \hline

\end{tabular}
}
\caption{Summary of the open-access input data and generate output volumes for the four scenarios investigated. Further information about each dataset can be found in the Supporting information }
\label{tab:Scenarios}
\end{table}

\begin{figure}[H]
    \centering
    \includegraphics[width=1\textwidth]{ 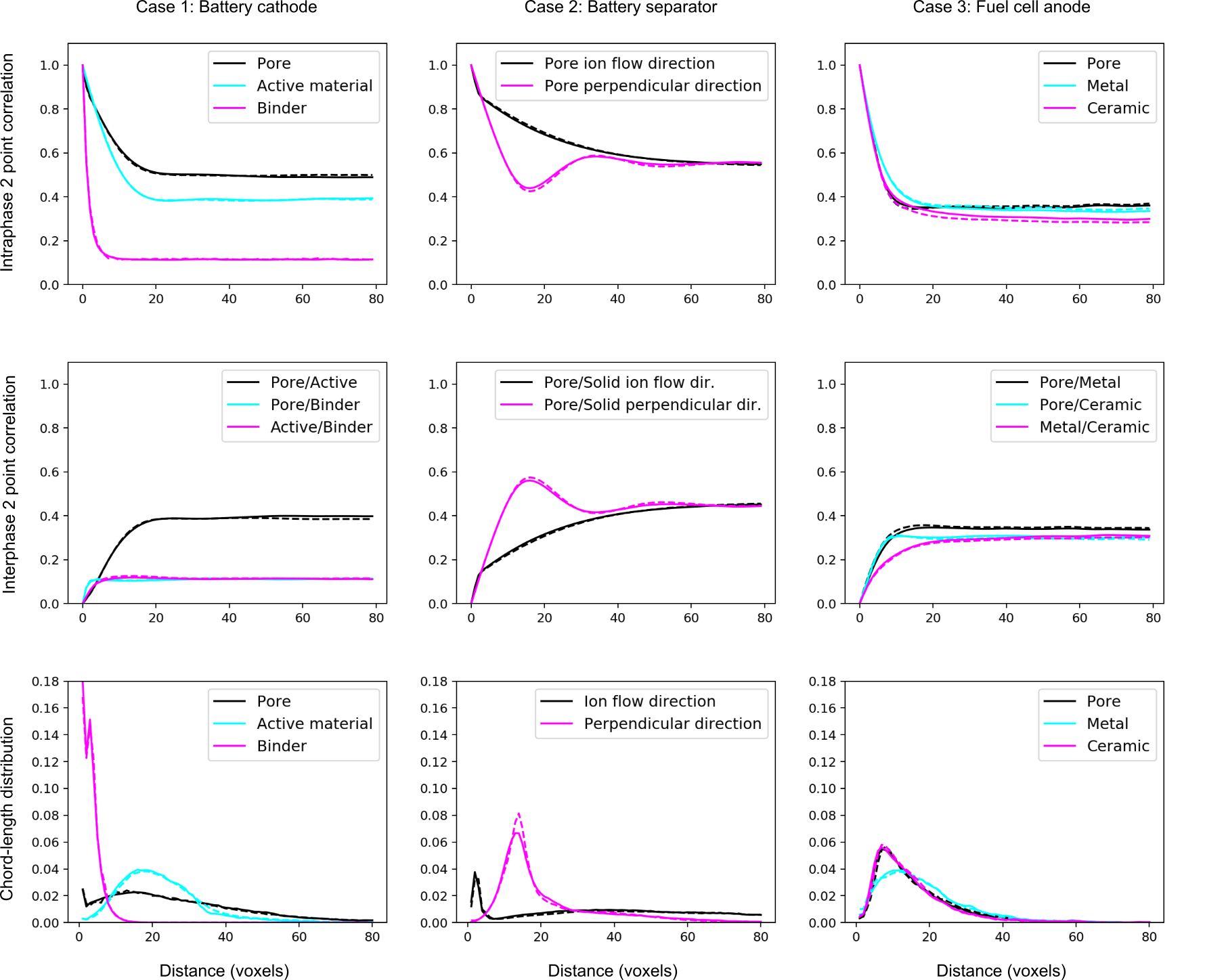}
    \caption{Further statistics comparing the generated volumes to the original volumes for the first three case studies. In all plots, the continuous line corresponds to the original volume and the dashed line corresponds to the generated super-res volume. The three columns refer to the first three case studies accordingly. The first row shows the calculation of the intraphase 2-point correlation function, which calculates the probability of reaching the same phase if moving $d$ voxels in a certain direction. The second row shows the calculation of the interphase 2-point correlation function, which is similar to the intraphase function, but calculates the probability of reaching a different phase if moving $d$ voxels in a certain direction, for $d\in \{0,\dots,80\}$. Notice that for the 2-phase separator material, the interphase function is just the complementary probability of the intraphase function. The last row shows the chord-length distribution, which is the probability of a voxel to be part of a $d$-length chord of the same phase in a certain direction. The distance 80 limit was chosen since the functions converged from this point onwards. For isotropic materials case studies 1 and 3, only one direction is shown since similar results were obtained for the other directions. For the 2-phase anisotropic case study 2, both the ion-flow direction and the perpendicular `through` the lamellae sheets direction are shown.}
    \label{fig:SI_2pc_cld}
\end{figure}

\begin{figure}[H]
    \centering
    \includegraphics[width=1\textwidth]{ 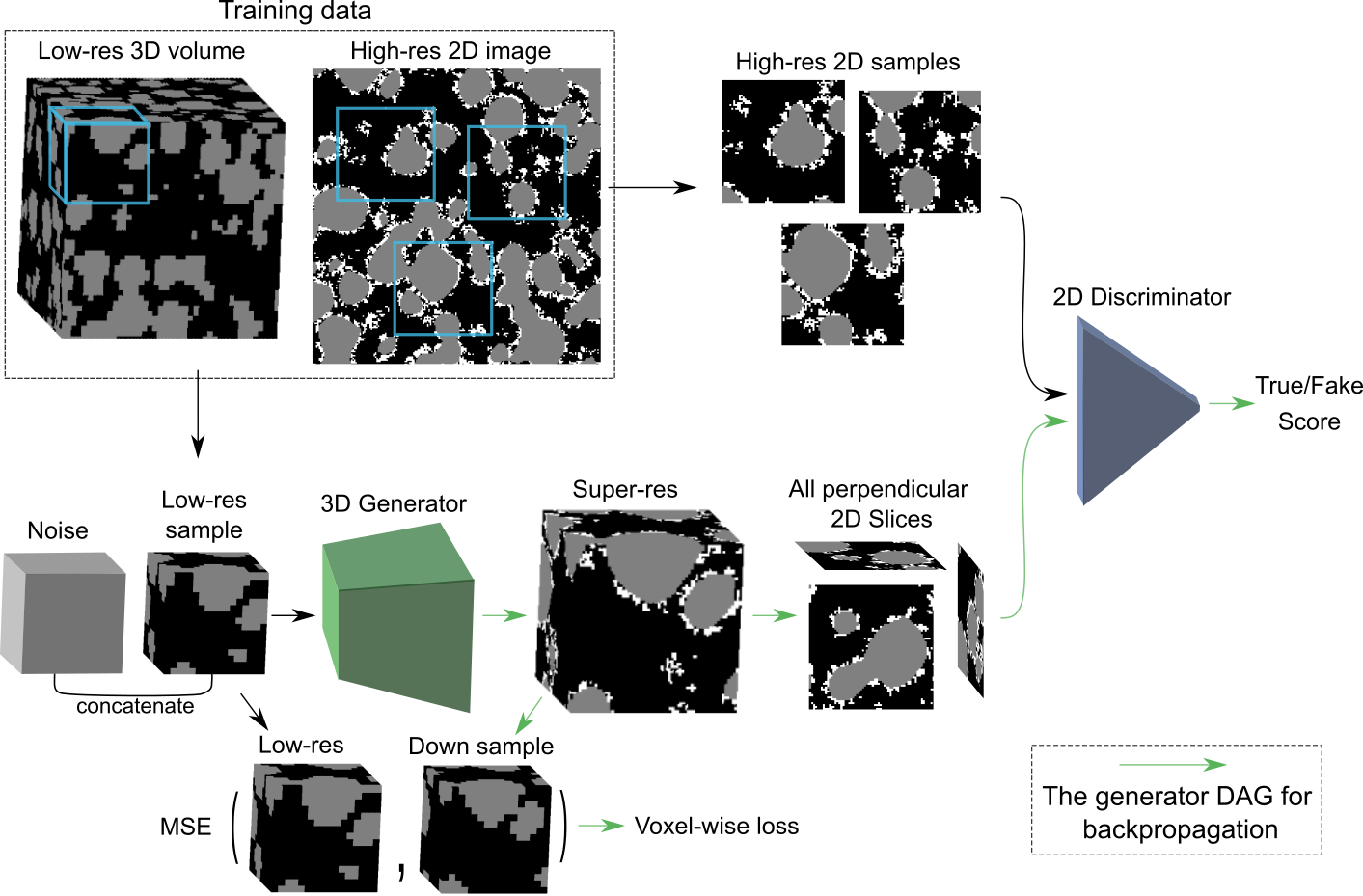}
    \caption{Model training diagram. Gaussian blurring is part of the down sampling as outlined in the Methods section.}
    \label{fig:training_outline}
\end{figure}

\begin{figure}[H]
    \centering
    \includegraphics[width=1\textwidth]{ 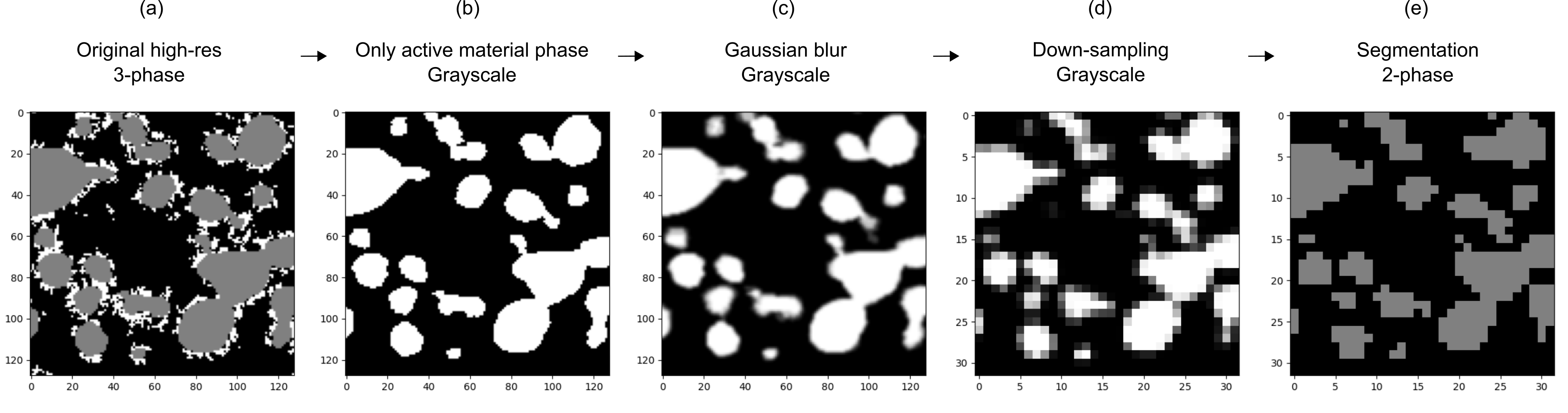}
    \caption{The process of creating the low-res input data volume from the high-res volume for case study 1. For better visualisation, a random $128^2$ sample (later $32^2$ after down-sampling) from a random slice was chosen from the volume. First, (a) shows a sample from the original high-res 3-phase volume. Then, (b) only the active material phase is chosen for blurring and down-sampling since the binder phase cannot be detected in low resolution using XCT. (b) is converted to gray-scale to (c) apply a gaussian blurring kernel and down-sampling (d) to approximately describe the physics limitation of the imaging technique. The gray-scale volume is then converted to (e) a 2-phase volume created by using a 0.5 threshold, mimicking a procedure made by researchers to obtain a segmented volume of the material.}
    \label{fig:creation_of_low_res_data}
\end{figure}

\begin{figure}[H]
    \centering
    \includegraphics[width=1\textwidth]{ 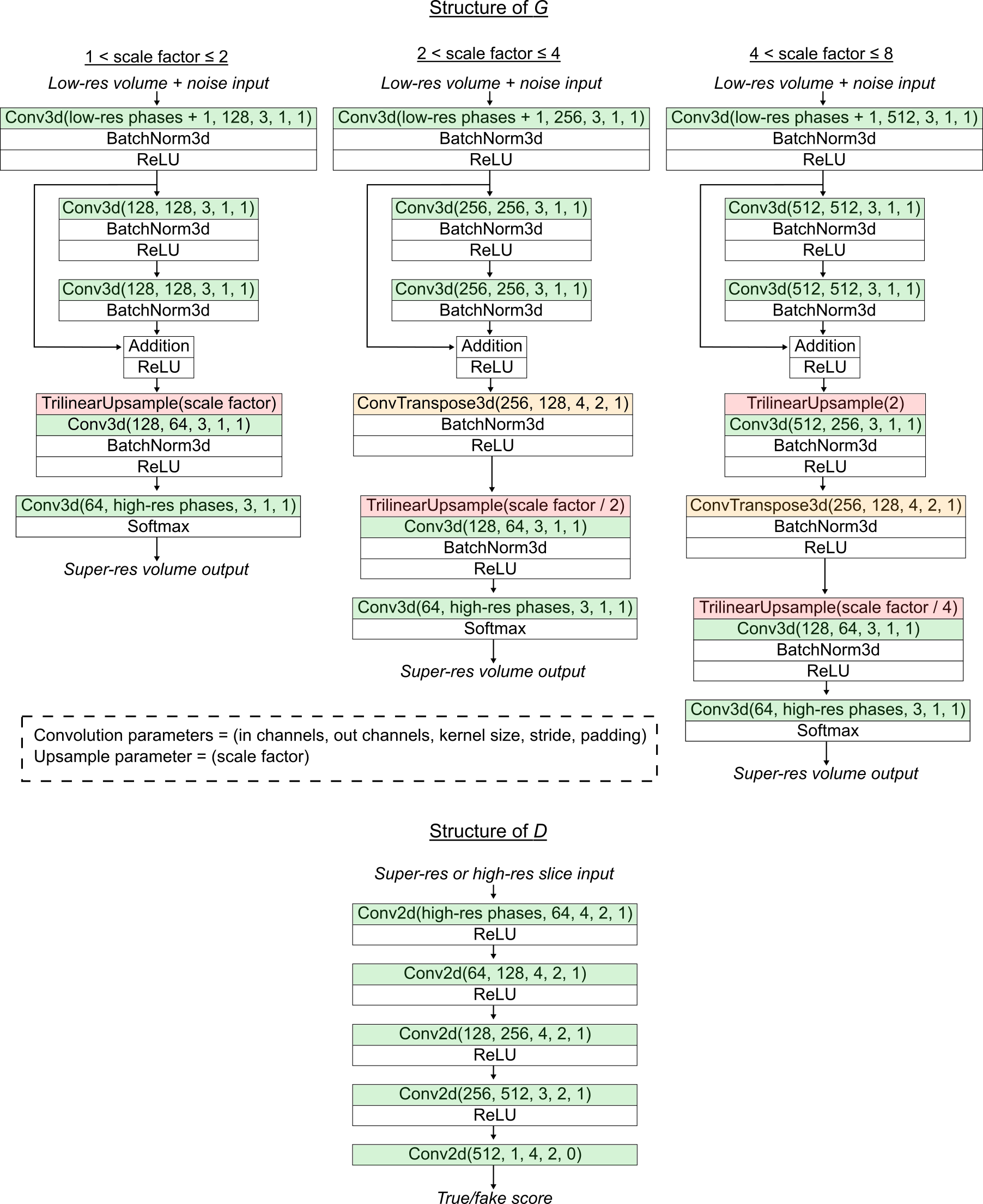}
    \caption{The structure of the generator ($G$) and the discriminator ($D$) in the GAN architecture. $G$'s structure is dependent on the scale factor used, and can have 3 different structures. This design enables the same output of $64^3$ cubes from different inputs, that are dependent on the scale factor. For example, low-res input with a scale factor of $2$ is a $32^3$ cube and a low-res input with a scale factor of $4$ is a $16^3$ cube and both will output a $64^3$ super-res cube. }
    \label{fig:g_structure}
\end{figure}

\begin{figure}[H]
    \centering
    \includegraphics[width=1\textwidth]{ 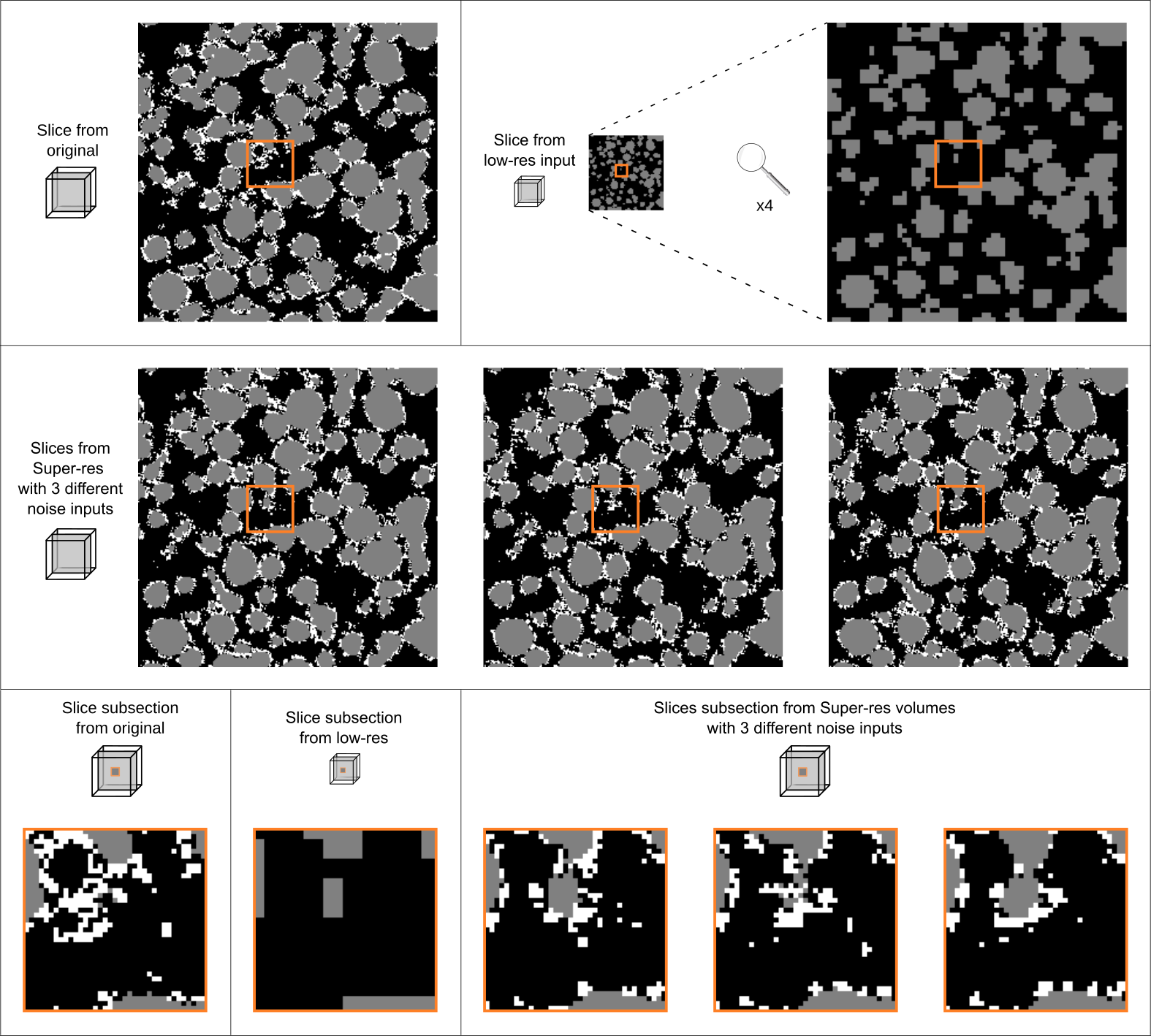}
    \caption{Presentation of different super-res volumes generated by the model of case study 1 using 3 different random noise inputs. The structure of this figure is similar to Figure \ref{fig:SEM_different_noise_seeds} with the addition of a slice from the original 3D high-res volume that the low-res volume that was derived from.}
    \label{fig:NMC_different_noise_seeds}
\end{figure}

\begin{figure}[H]
    \centering
    \includegraphics[width=1\textwidth]{ 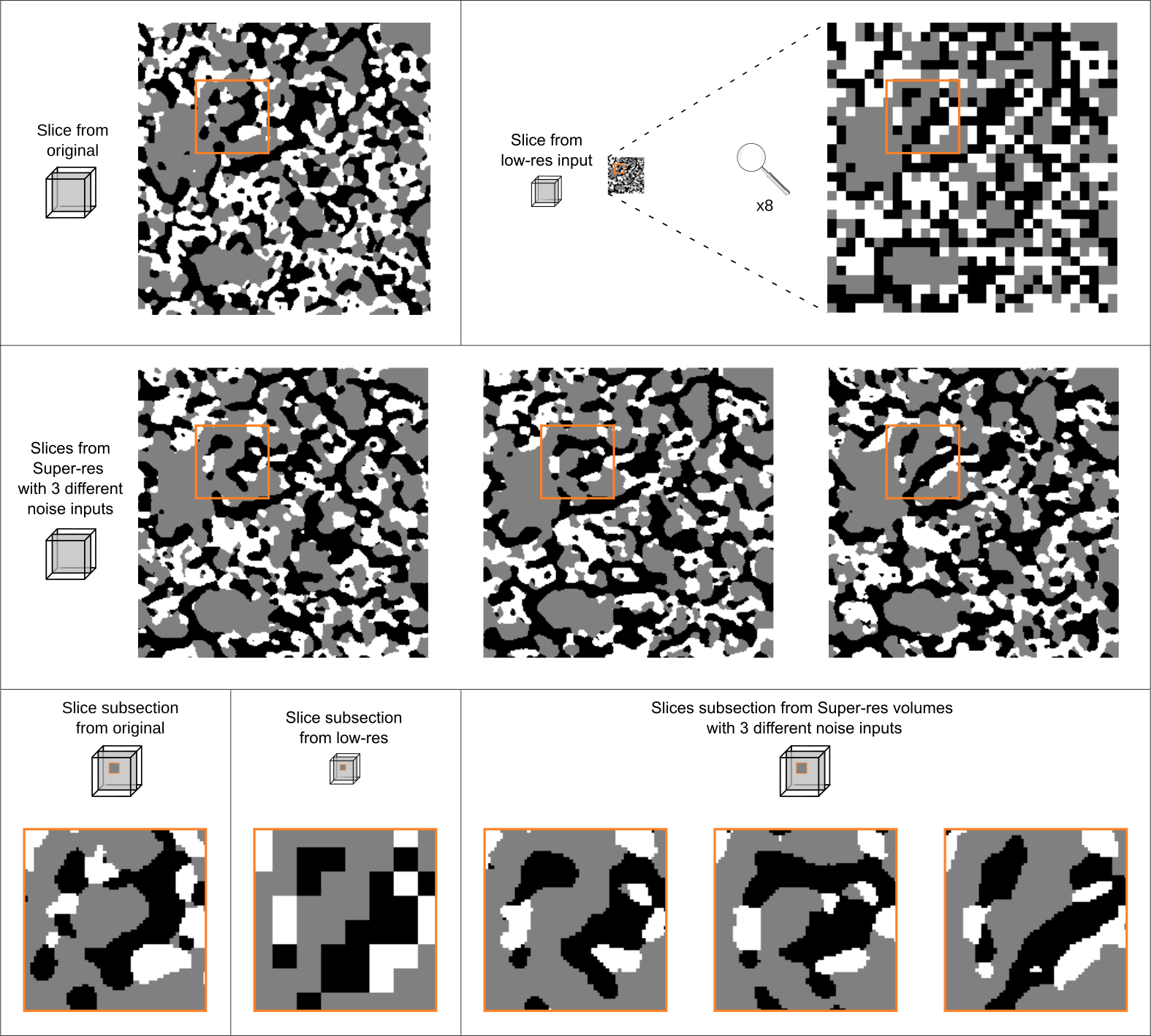}
    \caption{Presentation of different super-res volumes generated by the model of case study 3 with scale factor of 8 using 3 different random noise inputs. The structure of this figure is similar to Figure \ref{fig:NMC_different_noise_seeds}. \\\hspace{\textwidth}
    In this case of high scale factor, one voxel of the low-res input should represent $8^3=512$ voxels of the original volume. This difference in orders of magnitude naturally leads to a large loss of information when interpolating from the original volume to the low-res volume. As a result, there are a huge number of different reconstruction options for the super-res volume, all of which will be statistically similar to the original volume.}
    \label{fig:SOFC_x8_different_noises}
\end{figure}

\begin{figure}[H]
    \centering
    \includegraphics[width=1\textwidth]{ 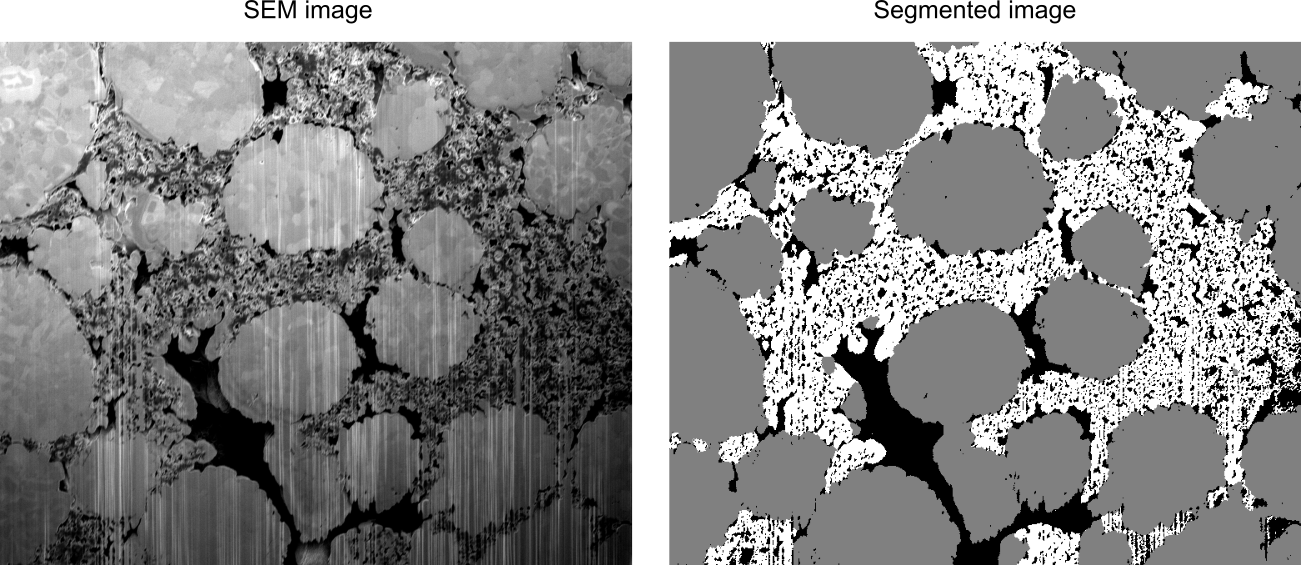}
    \caption{Segmentation of the 2D SEM image used in case study 4. The segmentation from greyscale to three-phase image was performed using the Trainable Weka Segmentation software package \cite{arganda2017trainable} , accompanied by some manual finishes to account for imaging artefacts. The ``curtaining" effect is visible as vertical streaks in the lower right-hand corner of the image. This caused Weka to mislabel some of the active material phase as binder, which we corrected manually. The ``pore-backs" effect caused some regions that perhaps should have been labelled as pore, to be labelled as binder instead, which we also adjusted manually. The total number of manually adjusted pixels represented less than 1\%of the image. The different segmented phases are pore (black), active material (grey) and binder (white). The original image is taken from an open source dataset \cite{usseglio2018resolving}.}
    \label{fig:sem_segmentation}
\end{figure}

\begin{figure}[H]
    \centering
    \includegraphics[width=1\textwidth]{ 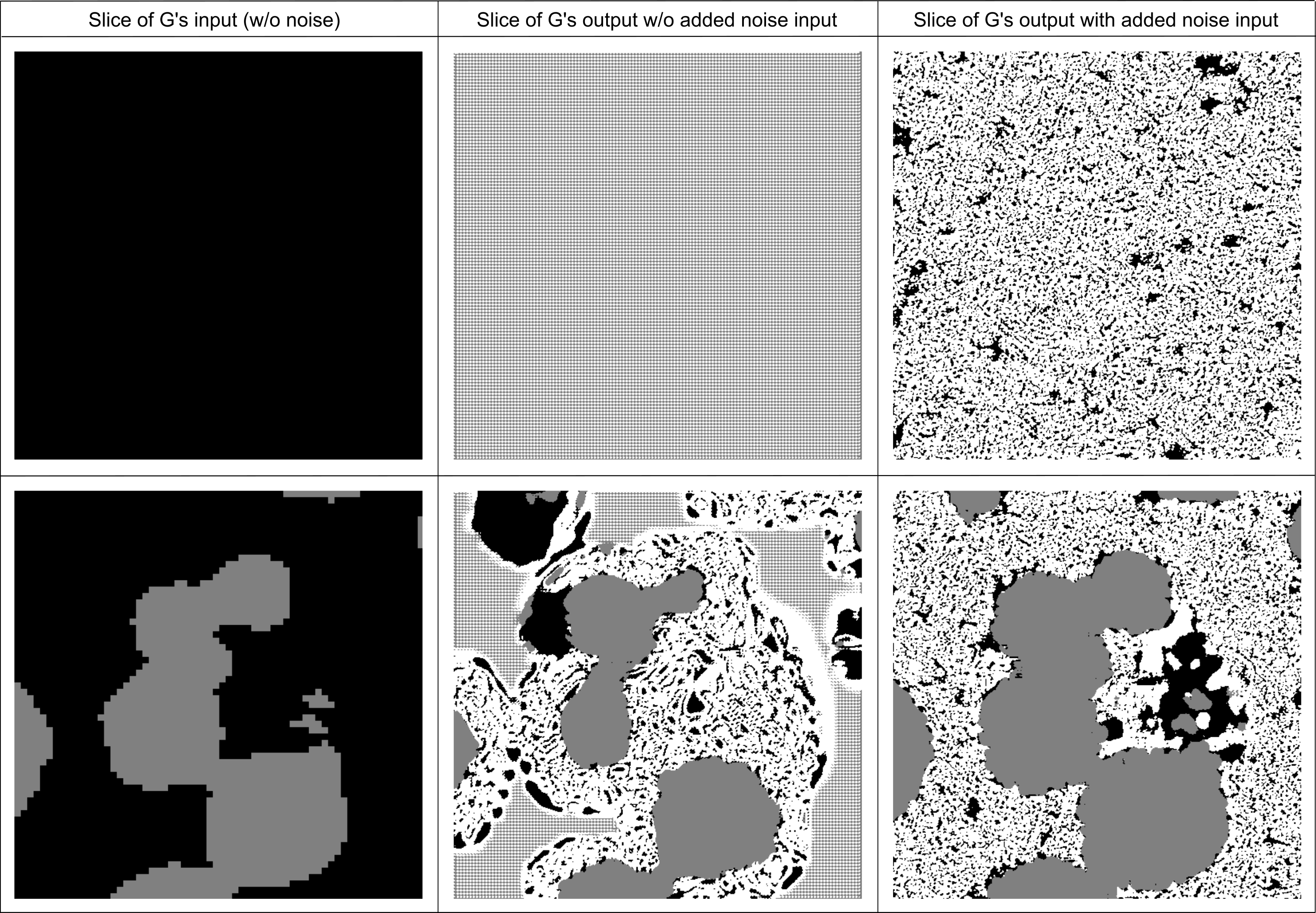}
    \caption{The importance of concatenating noise to the low-res input of $G$. A comparison is made between models that are trained with and without the concatenated input noise. Left column shows one slice of the low-res volume input to $G$, top left is from an all pore volume input and bottom left is from a volume that contains active material. The middle and right columns show slices of the outputs from models trained without (middle) and with (right) concatenated noise inputs. Respectably, the top row are slices of outputs from the all-pore volume (top left) and the bottom row are slices of outputs from the volume that contains active material (bottom left).}
    \label{fig:importance_of_noise}
\end{figure}

\begin{figure}[H]
    \centering
    \includegraphics[width=1\textwidth]{ 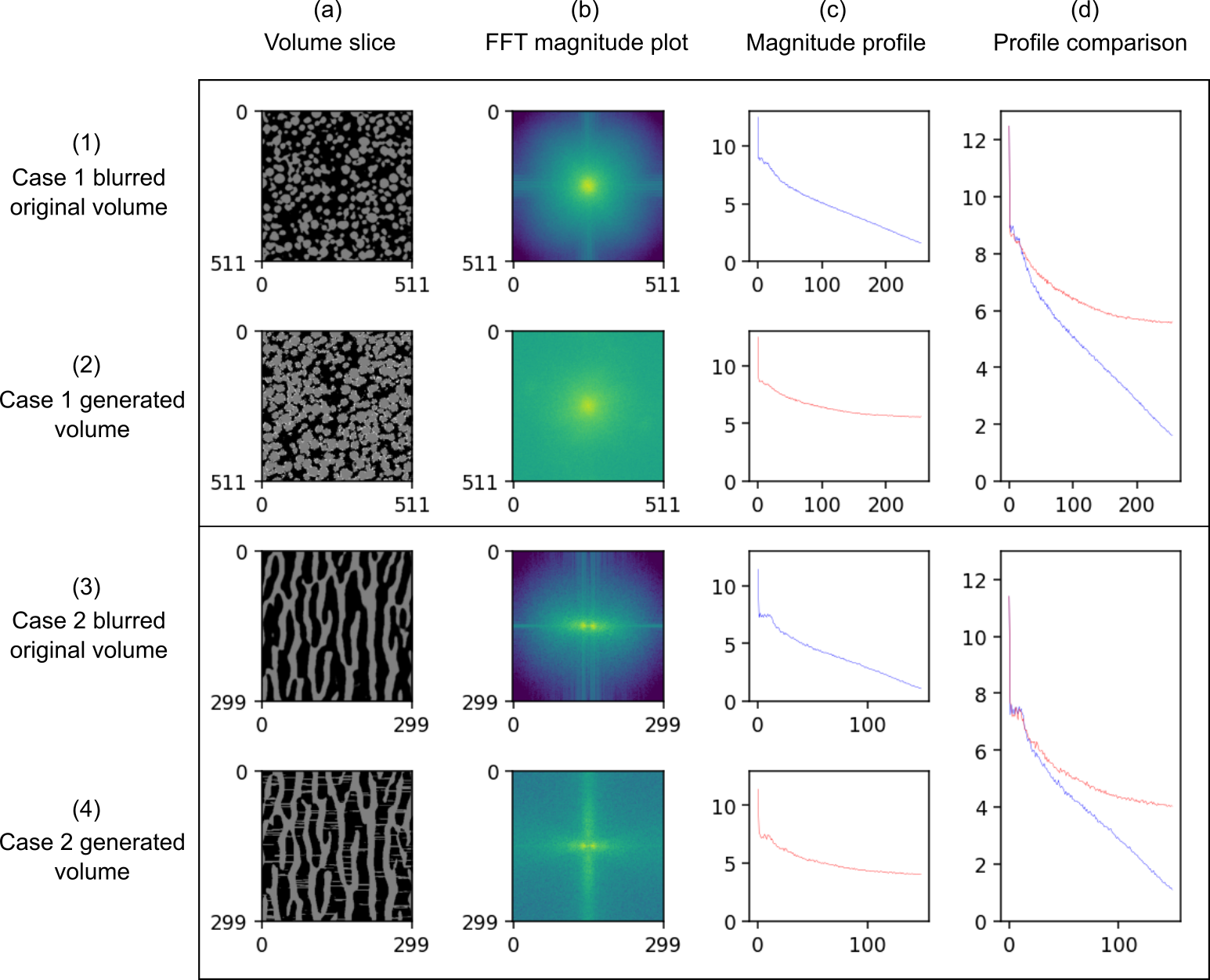}
    \caption{FFT magnitude plots comparison between the blurred original volumes and the generated super-res images made on the first two case studies. (a) The left column shows a slice of the respective volumes in the same location; column (b) shows the (centrally shifted) logarithmic FFT magnitude plots, while column (c) shows the plots magnitude profiles from the center towards higher frequencies by calculating the mean magnitude over receding rings around the center. Column (d) shows a comparison for each case study of the different magnitude plots. The profile comparison clearly shows a shifted elbow presence of higher-frequency information in the generated volumes, which indicates a higher resolution. For the n-phase image in the first case study, which is not grey-scale, the map is a mean of the FFT magnitude maps for the different phase maps (In this case the active material map and binder phase map). The blurred original volumes were chosen over the low-res input volumes since these images are in the same voxel-size as the generated volumes, for a simpler comparison. The low-res input, which is a result of down-sampling and segmentation of the blurred volume, contain less high fidelity information than the blurred volume.}
    \label{fig:fft_magnitude}
\end{figure}


	

%% file: main.bbl
\begin{thebibliography}{10}

\bibitem{arganda2017trainable}
I.~Arganda-Carreras, V.~Kaynig, C.~Rueden, K.~W. Eliceiri, J.~Schindelin,
  A.~Cardona, and H.~Sebastian~Seung.
\newblock Trainable weka segmentation: a machine learning tool for microscopy
  pixel classification.
\newblock {\em Bioinformatics}, 33(15):2424--2426, 2017.

\bibitem{arjovsky2017wasserstein}
M.~Arjovsky, S.~Chintala, and L.~Bottou.
\newblock Wasserstein generative adversarial networks.
\newblock In {\em International conference on machine learning}, pages
  214--223. PMLR, 2017.

\bibitem{burnett2019completing}
T.~Burnett and P.~Withers.
\newblock Completing the picture through correlative characterization.
\newblock {\em Nature materials}, 18(10):1041--1049, 2019.

\bibitem{cooper2016taufactor}
S.~J. Cooper, A.~Bertei, P.~R. Shearing, J.~Kilner, and N.~P. Brandon.
\newblock Taufactor: An open-source application for calculating tortuosity
  factors from tomographic data.
\newblock {\em SoftwareX}, 5:203--210, 2016.

\bibitem{cooper2022methods}
S.~J. Cooper, S.~A. Roberts, Z.~Liu, and B.~Winiarski.
\newblock Methods—kintsugi imaging of battery electrodes: Distinguishing
  pores from the carbon binder domain using pt deposition.
\newblock {\em Journal of The Electrochemical Society}, 2022.

\bibitem{fan2021adversarial}
L.~Fan, Z.~Wang, Y.~Lu, and J.~Zhou.
\newblock An adversarial learning approach for super-resolution enhancement
  based on agcl@ ag nanoparticles in scanning electron microscopy images.
\newblock {\em Nanomaterials}, 11(12):3305, 2021.

\bibitem{finegan2016characterising}
D.~P. Finegan, S.~J. Cooper, B.~Tjaden, O.~O. Taiwo, J.~Gelb, G.~Hinds, D.~J.
  Brett, and P.~R. Shearing.
\newblock Characterising the structural properties of polymer separators for
  lithium-ion batteries in 3d using phase contrast x-ray microscopy.
\newblock {\em Journal of Power Sources}, 333:184--192, 2016.

\bibitem{furat2022super}
O.~Furat, D.~P. Finegan, Z.~Yang, T.~Kirstein, K.~Smith, and V.~Schmidt.
\newblock Super-resolving microscopy images of li-ion electrodes for
  fine-feature quantification using generative adversarial networks.
\newblock {\em npj Computational Materials}, 8(1):1--11, 2022.

\bibitem{gatys2016image}
L.~A. Gatys, A.~S. Ecker, and M.~Bethge.
\newblock Image style transfer using convolutional neural networks.
\newblock In {\em Proceedings of the IEEE conference on computer vision and
  pattern recognition}, pages 2414--2423, 2016.

\bibitem{Gayon-Lombardo2020}
A.~Gayon-Lombardo, L.~Mosser, N.~P. Brandon, and S.~J. Cooper.
\newblock {Pores for thought: Generative adversarial networks for the
  stochastic reconstruction of 3D multi-phase electrode microstructures with
  periodic boundaries}.
\newblock {\em npj Computational Materials}, 6(1):82, feb 2020.

\bibitem{goodfellow2014generative}
I.~J. Goodfellow, J.~Pouget-Abadie, M.~Mirza, B.~Xu, D.~Warde-Farley, S.~Ozair,
  A.~Courville, and Y.~Bengio.
\newblock Generative adversarial networks.
\newblock {\em arXiv preprint arXiv:1406.2661}, 2014.

\bibitem{hsu2018mesoscale}
T.~Hsu, W.~K. Epting, R.~Mahbub, N.~T. Nuhfer, S.~Bhattacharya, Y.~Lei, H.~M.
  Miller, P.~R. Ohodnicki, K.~R. Gerdes, H.~W. Abernathy, et~al.
\newblock Mesoscale characterization of local property distributions in
  heterogeneous electrodes.
\newblock {\em Journal of Power Sources}, 386:1--9, 2018.

\bibitem{jackson2022deep}
S.~J. Jackson, Y.~Niu, S.~Manoorkar, P.~Mostaghimi, and R.~T. Armstrong.
\newblock Deep learning of multiresolution x-ray micro-computed-tomography
  images for multiscale modeling.
\newblock {\em Physical Review Applied}, 17(5):054046, 2022.

\bibitem{Jones2014}
H.~G. Jones, K.~P. Mingard, and D.~C. Cox.
\newblock {Investigation of slice thickness and shape milled by a focused ion
  beam for three-dimensional reconstruction of microstructures}.
\newblock {\em Ultramicroscopy}, 139(0):20--28, 2014.

\bibitem{jung2021super}
J.~Jung, J.~Na, H.~K. Park, J.~M. Park, G.~Kim, S.~Lee, and H.~S. Kim.
\newblock Super-resolving material microstructure image via deep learning for
  microstructure characterization and mechanical behavior analysis.
\newblock {\em npj Computational Materials}, 7(1):1--11, 2021.

\bibitem{karras2020analyzing}
T.~Karras, S.~Laine, M.~Aittala, J.~Hellsten, J.~Lehtinen, and T.~Aila.
\newblock Analyzing and improving the image quality of stylegan.
\newblock In {\em Proceedings of the IEEE/CVF conference on computer vision and
  pattern recognition}, pages 8110--8119, 2020.

\bibitem{kench2021generating}
S.~Kench and S.~J. Cooper.
\newblock Generating three-dimensional structures from a two-dimensional slice
  with generative adversarial network-based dimensionality expansion.
\newblock {\em Nature Machine Intelligence}, pages 1--7, 2021.

\bibitem{Kroll2021}
M.~Kroll, S.~L. Karstens, M.~Cronau, A.~H{\"{o}}ltzel, S.~Schlabach, N.~Nobel,
  C.~Redenbach, B.~Roling, and U.~Tallarek.
\newblock {Three‐Phase Reconstruction Reveals How the Microscopic Structure
  of the Carbon‐Binder Domain Affects Ion Transport in Lithium‐Ion
  Batteries}.
\newblock {\em Batteries \& Supercaps}, may 2021.

\bibitem{Lagadec2016}
M.~F. Lagadec, M.~Ebner, R.~Zahn, and V.~Wood.
\newblock Communication{\textemdash}technique for visualization and
  quantification of lithium-ion battery separator microstructure.
\newblock {\em Journal of The Electrochemical Society}, 163(6):A992--A994,
  2016.

\bibitem{ledig2017photo}
C.~Ledig, L.~Theis, F.~Husz{\'a}r, J.~Caballero, A.~Cunningham, A.~Acosta,
  A.~Aitken, A.~Tejani, J.~Totz, Z.~Wang, et~al.
\newblock Photo-realistic single image super-resolution using a generative
  adversarial network.
\newblock In {\em Proceedings of the IEEE conference on computer vision and
  pattern recognition}, pages 4681--4690, 2017.

\bibitem{lemmens2011fib}
H.~Lemmens, A.~Butcher, and P.~Botha.
\newblock Fib/sem and sem/edx: a new dawn for the sem in the core lab?
\newblock {\em Petrophysics-The SPWLA Journal of Formation Evaluation and
  Reservoir Description}, 52(06):452--456, 2011.

\bibitem{Lu2020a}
X.~Lu, A.~Bertei, D.~P. Finegan, C.~Tan, S.~R. Daemi, J.~S. Weaving, K.~B.
  O'Regan, T.~M. Heenan, G.~Hinds, E.~Kendrick, D.~J. Brett, and P.~R.
  Shearing.
\newblock {3D microstructure design of lithium-ion battery electrodes assisted
  by X-ray nano-computed tomography and modelling}.
\newblock {\em Nature Communications}, 11(1):1--13, apr 2020.

\bibitem{mescheder2018training}
L.~Mescheder, A.~Geiger, and S.~Nowozin.
\newblock Which training methods for gans do actually converge?
\newblock In {\em International conference on machine learning}, pages
  3481--3490. PMLR, 2018.

\bibitem{Moroni2020}
R.~Moroni and S.~Thiele.
\newblock {FIB/SEM tomography segmentation by optical flow estimation}.
\newblock {\em Ultramicroscopy}, 219:113090, dec 2020.

\bibitem{niu2021towards}
Z.~Niu, V.~J. Pinfield, B.~Wu, H.~Wang, K.~Jiao, D.~Y. Leung, and J.~Xuan.
\newblock Towards the digitalisation of porous energy materials: evolution of
  digital approaches for microstructural design.
\newblock {\em Energy \& Environmental Science}, 14(5):2549--2576, 2021.

\bibitem{paszke2019pytorch}
A.~Paszke, S.~Gross, F.~Massa, A.~Lerer, J.~Bradbury, G.~Chanan, T.~Killeen,
  Z.~Lin, N.~Gimelshein, L.~Antiga, et~al.
\newblock Pytorch: An imperative style, high-performance deep learning library.
\newblock {\em Advances in neural information processing systems}, 32, 2019.

\bibitem{Pietsch2018}
P.~Pietsch, M.~Ebner, F.~Marone, M.~Stampanoni, and V.~Wood.
\newblock Determining the uncertainty in microstructural parameters extracted
  from tomographic data.
\newblock {\em Sustainable Energy Fuels}, 2:598--605, 2018.

\bibitem{stoyan2013stochastic}
D.~Stoyan, W.~S. Kendall, S.~N. Chiu, and J.~Mecke.
\newblock {\em Stochastic geometry and its applications}.
\newblock John Wiley \& Sons, 2013.

\bibitem{usseglio2018resolving}
F.~L. Usseglio-Viretta, A.~Colclasure, A.~N. Mistry, K.~P.~Y. Claver,
  F.~Pouraghajan, D.~P. Finegan, T.~M. Heenan, D.~Abraham, P.~P. Mukherjee,
  D.~Wheeler, et~al.
\newblock Resolving the discrepancy in tortuosity factor estimation for li-ion
  battery electrodes through micro-macro modeling and experiment.
\newblock {\em Journal of The Electrochemical Society}, 165(14):A3403, 2018.

\bibitem{wang2020deep}
Z.~Wang, J.~Chen, and S.~C. Hoi.
\newblock Deep learning for image super-resolution: A survey.
\newblock {\em IEEE transactions on pattern analysis and machine intelligence},
  2020.

\bibitem{xu2020microstructure}
H.~Xu, F.~Usseglio-Viretta, S.~Kench, S.~J. Cooper, and D.~P. Finegan.
\newblock Microstructure reconstruction of battery polymer separators by fusing
  2d and 3d image data for transport property analysis.
\newblock {\em Journal of Power Sources}, 480:229101, 2020.

\end{thebibliography}
